\documentclass[10pt,journal,compsoc]{IEEEtran}
\usepackage[numbers]{natbib}
\usepackage{hyperref}
\usepackage{url}
\usepackage{booktabs}       
\usepackage{amsfonts}       
\usepackage{nicefrac}       
\usepackage{microtype}      
\usepackage{xcolor}         

\usepackage{graphicx}
\usepackage{amsthm}
\usepackage{amssymb}
\usepackage{amsmath,amsfonts}
\usepackage{mathtools}
\usepackage[noend,ruled,vlined,linesnumbered]{algorithm2e}
\usepackage{dirtytalk}
\usepackage{subcaption}
\usepackage{wrapfig}
\usepackage{stfloats}
\usepackage[section]{placeins}
\usepackage{color}
\usepackage{soul}
\usepackage{ulem}
\usepackage{xr}

\theoremstyle{definition}
\newtheorem{definition}{Definition}
\newtheorem{assumption}{Assumption}
\newtheorem{method}{Method}
\newtheorem{lemma}{Lemma}

\renewcommand{\citet}[1]{\citep{#1}}

\ifCLASSOPTIONcompsoc
  \usepackage[nocompress]{cite}
\else
  \usepackage{cite}
\fi
\ifCLASSINFOpdf
\else
\fi
\begin{document}
%
\title{The Fragility of Noise Estimation in Kalman Filter:\\Optimization Can Handle Model-Misspecification}
%
%
%
%

\author{Ido~Greenberg,
        Shie~Mannor
        and~Netanel~Yannay
\IEEEcompsocitemizethanks{\IEEEcompsocthanksitem Greenberg is with the Technion, Israel. 
E-mail: gido@campus.technion.ac.il
\IEEEcompsocthanksitem Mannor is with the Technion, Israel, and with Nvidia Research.}
\thanks{This work was partially funded by the Israel Science Foundation under ISF grant number 2199/20.}
}

\markboth{}%
{Greenberg \MakeLowercase{\textit{et al.}}: The Fragility of Noise Estimation in Kalman Filter}
%



\IEEEtitleabstractindextext{%
\begin{abstract}
The Kalman Filter (KF) parameters are traditionally determined by noise estimation, since under the KF assumptions, the state prediction errors are minimized when the parameters correspond to the noise covariance.
However, noise estimation remains the gold-standard regardless of the assumptions -- even when it is not equivalent to errors minimization. 
We demonstrate that even seemingly simple problems may include multiple assumptions violations -- which are sometimes hard to even notice.
We show theoretically and empirically that even a minor violation may largely shift the optimal parameters.
We propose a gradient-based method along with the Cholesky parameterization to explicitly optimize the state prediction errors. 
We show consistent improvement over noise estimation in tens of experiments in 3 different domains.
Finally, we demonstrate that optimization makes the KF competitive with an LSTM model -- even in non linear problems.
We publish our experiments in \href{https://github.com/ido90/UsingKalmanFilterTheRightWay}{\underline{Github}} and the Optimized-KF (OKF) in \href{https://pypi.org/project/Optimized-Kalman-Filter/}{\underline{PyPI}}.
\end{abstract}

\begin{IEEEkeywords}
Kalman Filter, noise estimation, optimization, gradient descent, Cholesky parameterization.
\end{IEEEkeywords}}

\maketitle

\IEEEdisplaynontitleabstractindextext

%
\IEEEpeerreviewmaketitle

\IEEEraisesectionheading{\section{Introduction}\label{sec:intro}}

\IEEEPARstart{T}{he} Kalman Filter (KF)~\citep{KF} is a celebrated method for linear filtering and prediction, with applications in many fields including tracking, guidance, navigation and control~\citep{KF_practical_guide,tracking_and_navigation}.
Due to its simplicity and robustness, it remains highly popular -- with over 10,000 citations in the last 5 years alone~\citep{KF_citations} -- despite the rise of many non-linear sequential prediction models (e.g., recurrent neural networks).
The KF relies on the following model for a dynamic system:
\begin{align}
\label{eq:KF_model}
\begin{split}
    X_{t+1} &= F_tX_t + \omega_t \qquad (\omega_t\sim N(0,Q)) \\ Z_{t} &= H_tX_t + \nu_t \qquad (\nu_t\sim N(0,R)) ,
\end{split}
\end{align}
where $X_t$ is the state of the system at time $t$ (whose estimation is usually the goal), and its dynamics are modeled by the linear operator $F_t$ up to the random noise $\omega_t$ with covariance $Q$; and $Z_t$ is the observation, which is modeled by the linear operator $H_t$ up to the noise $\nu_t$ with covariance $R$. When the operators $F_t,H_t$ are not assumed to depend on time, the notation may be simplified to $F,H$.

To use the KF, one must determine the noise parameters $Q,R$.
The filtering errors (i.e., estimation errors of the states $\{X_t\}$) are minimized when $Q$ and $R$ correspond to the true covariance matrices of the noise~\citep{KF_fresh_look}. Thus, these parameters are usually determined by noise estimation.
In absence of system state data $\{x_t\}$ (the "ground truth"), many methods have been suggested to determine $Q$ and $R$ from observations data $\{z_t\}$ alone \citep{Thrun_trainingKF,ALS,cov_estimation_varying_processes,measurement_noise_recommendation}.
When ground-truth data \textit{is} available, however, noise estimation trivially reduces to calculation of the sample covariance matrices~\citep{TutorialKF}:
\begin{align}
\label{eq:noise_estimation}
\begin{split}
    \hat{R}&\coloneqq Cov(\{z_t-H_tx_t\}_t) \\ \hat{Q}&\coloneqq Cov(\{x_{t+1}-F_tx_t\}_t)
\end{split}
\end{align}
Our work focuses on such problems where the ground-truth is available for learning (but not for inference after the learning, of course), motivated by a real-world Doppler radar problem.


\textbf{Filtering when the KF assumptions break:}
It is well-known that the KF yields optimal state-estimations only under a restrictive set of assumptions: known and linear dynamics and observation models ($F_t,H_t$), with i.i.d and normally-distributed noises ($\{\omega_t\},\{\nu_t\}$)~\citep{KF_fresh_look}.
When the assumptions break, alternative models are sometimes considered -- from the Extended KF (EKF)~\citep{KF_theory} to more recent neural network models (NNs), which can be optimized with respect to the filtering errors.
However, when the KF \textit{is} used in presence of ground-truth data, it is usually tuned by the gold-standard of noise estimation (if not heuristically) -- regardless of the assumptions. Even if noise estimation is not equivalent anymore to errors minimization, the possibility of an explicit and systematic optimization \textit{without} switching the whole model (e.g., to a NN) is usually not considered.


Non-optimized KFs are particularly common in the literature of non-linear filtering, where optimized learning models such as NNs are often compared to the non-optimized baseline of a KF (or EKF).
In many such works, the KF parameters are tuned by noise estimation \citep{navigation_using_RNN,ANN_vs_EKF,KalmanNet}; by heuristics \citep{learning_in_indoor_navigation,pose_estimation,KF_with_ANN}; or are simply ignored \citep{ManeuveringTargetTracking2,bai2020neuron,RNN_EKF}.
\citet{KF_vs_NN_for_batteries} explicitly discusses the sensitivity of EKF to its noise model, and suggests the solution of a NN with supervised learning -- without ever considering the same supervised learning for the EKF itself.
Indeed, even when ground-truth data is \textit{already used} for supervised learning of other models, the KF is still not optimized in a similar manner.

Optimization is sometimes used to tune the KF from observations alone~\citep{BoydOptimization}, but as demonstrated above, it is hardly considered whenever noise estimation from ground-truth data is possible.
As stated by \citet{ALS}, \say{\textit{the more systematic and preferable approach to determine the filter gain is to estimate the covariances from data}}.
In this work we disprove this statement theoretically and empirically.

To that end, we propose a gradient-based method for Optimization of KF (\textit{OKF}) with respect to the filtering errors.
We use the Cholesky parameterization to preserve the symmetric and positive-definite (SPD) structure of $Q$ and $R$.


\textbf{Main results:}
We demonstrate that OKF improves the KF errors versus noise estimation consistently -- over different KF variants, over different violations of KF assumptions, over different domains (tracking from radar, video or lidar), over small and large training datasets, and even under distributional shifts between train and test datasets.

In fact, OKF beats not only noise \textit{estimation}, but even the \textit{true} covariance of the noise ("oracle" noise estimation).
That is, the limitations of noise estimation come not from inaccuracy but from discrepancy of objectives. 
Indeed, Appendices~\ref{sec:toy_analysis} and \ref{sec:theory_noniid} theoretically analyze the discrepancy between the noise covariances and the optimal parameters under minor assumptions violations -- violations which may sometimes be tricky to even notice.

We further show that seemingly small changes in the properties of the scenario may lead to major changes in the desired design of the KF, e.g., whether to use a KF or an EKF. 
In certain cases, the design choices are easy to overlook (e.g., Cartesian vs. spherical coordinates), and are not trivial to make even if they are noticed.
OKF is demonstrated to improve the robustness to such design decisions, by shrinking the performance gaps between different variants of the KF.

Finally, we consider an extension of the KF based on LSTM, which is the key component in many SOTA algorithms for non-linear sequential prediction~\citep{process_prediction_review}.
In the non-linear radar problem, the LSTM provides a significant improvement over the KF -- similarly to the advantage of NNs in the works discussed above.
However, {\bf the KF becomes competitive with the LSTM once they are optimized similarly}:
the whole improvement comes from optimization and \textit{not} from the non-linear architecture.
This result joins recent works in the area of machine learning, that show how advanced algorithms often obtain their improvement mostly from implementation nuances~\citep{RL_reproducibility,implementation_matters,what_matters_in_RL}.

\textbf{Scope:}
Our setting includes training data with ground-truth system-states.
Ground-truth data is often available from controlled experiments, simulations or manual labeling.
This setting is common in the \textit{non-linear} filtering literature (see Section~\ref{sec:related_work}).
The \textit{KF} literature, however, often focuses on algorithms that do \textit{not} exploit ground-truth data, since exploitation of such data is considered solved by noise estimation (Eq.~\ref{eq:noise_estimation}).
Although noise estimation is indeed simple given such data, we show that it is often not the right task to address.

\textbf{Limitations:}
We justify the sub-optimality of noise estimation theoretically, and the advantage of our method empirically.
However, as discussed in Appendix~\ref{sec:GD_limitations}, theoretical guarantees for gradient-based optimization methods are limited, despite their impressive success in machine learning (which includes non-convex problems with hundreds of millions of parameters~\citep{bert}).

The optimization also loses the independent physical meaning of the KF parameters as representatives of the noise. Yet, as demonstrated in Appendices~\ref{sec:toy_analysis} and \ref{sec:theory_noniid}, the optimized parameters carry certain meaning about the \textit{effective} noise corresponding to the state estimation task.

\textbf{Contributions:} Our contributions are as follows:
\begin{enumerate}
    \item \textit{Why optimize?} We provide extensive theoretical and empirical analysis of the sub-optimality of the standard noise estimation method under KF model-misspecification. 
    \item \textit{How to optimize?} We introduce OKF -- a method for gradient-based optimization of KF errors from ground-truth data using Cholesky parameterization. While all the ingredients are well-known, to the best of our knowledge we are the first to put them together.
    We also provide an off-the-shelf implementation of OKF as a \href{https://pypi.org/project/Optimized-Kalman-Filter/}{\underline{PyPI}} package.
    \item An optimized KF is demonstrated to be competitive with LSTM -- even in non-linear problems. In particular, this indicates a methodological flaw in many works that compare \textit{optimized} models to a \textit{non-optimized} KF.
\end{enumerate}

The paper is organized as follows: Section~\ref{sec:preliminaries} reviews the KF.
Section~\ref{sec:SPD_optimization} introduces OKF.
Section~\ref{sec:OKF} motivates the KF optimization through a detailed case study.
Section~\ref{sec:NKF} presents an LSTM-based extension of the KF, which seems to reduce the errors compared to a KF -- but only until the KF is optimized.
Section~\ref{sec:more_experiments} tests OKF in further experimental domains.
Section~\ref{sec:related_work} discusses related works. 
In addition, Appendices~\ref{sec:toy_analysis} and \ref{sec:theory_noniid} present the complete theoretical analysis, and Appendices~\ref{sec:generalization}, \ref{sec:train_size} and \ref{sec:NKF_extended} provide extended experiments.

\section{Preliminaries: the Kalman Filter}
\label{sec:preliminaries}



\begin{figure}[!h]
  \begin{center}
    \includegraphics[width=\linewidth]{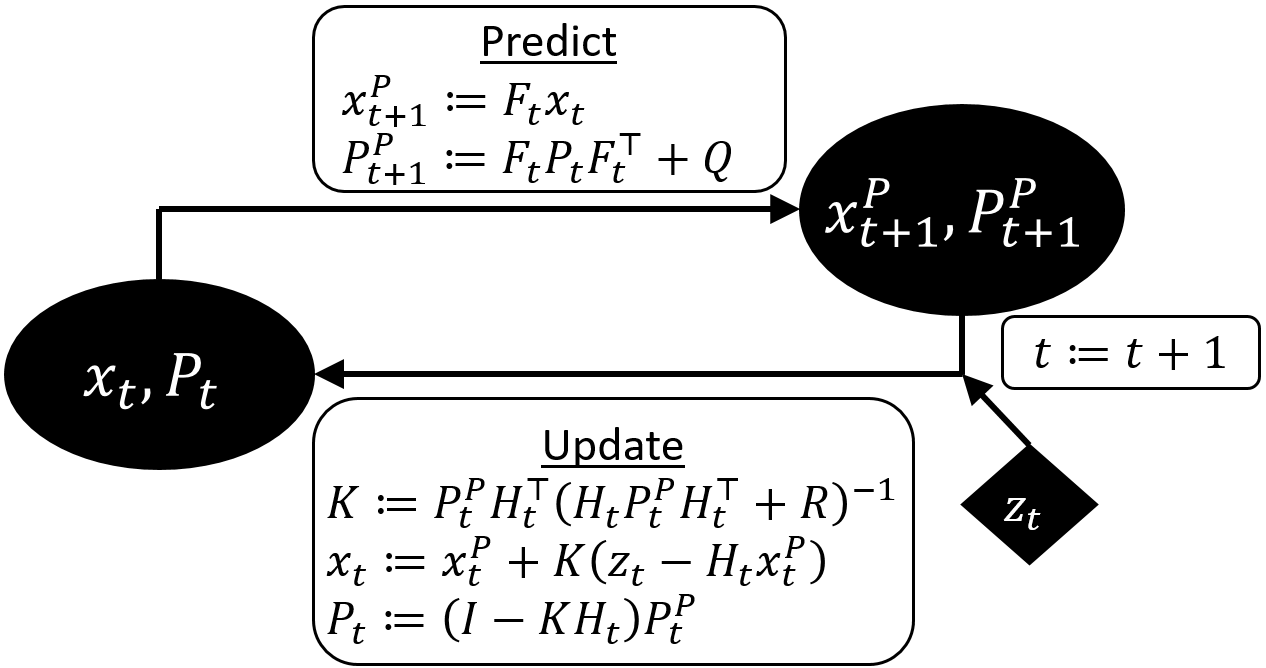}
  \end{center}
\caption{\small The KF algorithm. The prediction step is based on the motion model $F_t$ with noise $Q$, whereas the update step is based on the observation model $H_t$ with noise $R$.}
\label{fig:KF}
\end{figure}

The KF algorithm~\citep{KF,KF_fresh_look} relies on Eq.~\ref{eq:KF_model} for a dynamic system model.
It keeps an estimate of the state $X_t$, represented as the mean $x_t$ and covariance $P_t$ of a normal distribution. 
As shown in Figure~\ref{fig:KF}, it alternately predicts the next state using the dynamics model (\textit{prediction} step), and processes new information from incoming observations (\textit{update} or \textit{filtering} step).

The KF yields optimal state estimation -- but only under a restrictive set of assumptions~\citep{KF}, as specified in Assumption~\ref{def:KF_assumptions}.
Note that normality of the noise is excluded since it is not necessary for parameters optimality~\citep{KF_fresh_look}. 

\begin{assumption}[KF assumptions]
\label{def:KF_assumptions}
$F_t,H_t$ of Eq.~\ref{eq:KF_model} are known matrices that do not depend on the state (i.e., correspond to linear models); $\omega_t,\nu_t$ are i.i.d random variables with zero-mean and constant, known covariance matrices $Q,R$, respectively; and the initial state distribution is known. 
\end{assumption}


Violation of Assumption~\ref{def:KF_assumptions} can be partially handled by variations of the KF, such as the Extended KF (EKF)~\citep{KF_theory} which replaces the linear models $F_t,H_t$ with local linear approximations, and the Unscented KF (UKF)~\citep{UKF} which applies the filtering through sigma-points sampled from the estimated distribution. 

While $F_t$ and $H_t$ are usually determined based on domain knowledge, $Q$ and $R$ are often estimated from data as the covariances of the noise. As mentioned in Section~\ref{sec:intro}, this can be done using Eq.~\ref{eq:noise_estimation} if ground-truth data is available, or using more sophisticated methods otherwise~\citep{Mehra,ALS,seq_cov_estimation,measurement_noise_recommendation}.


\section{Kalman Filter Optimization} 
\label{sec:SPD_optimization}

The performance of a KF strongly depends on its parameters $Q$ and $R$~\citep{noise_cov_estimation}. These parameters are usually regarded as estimators of the noise covariance in dynamics and observations, respectively~\citep{TutorialKF}, and are estimated accordingly.
Although optimization has been suggested for the KF in the past~\citep{Thrun_trainingKF}, it was often viewed as a fallback solution~\citep{ALS}, for cases where direct estimation is not possible (e.g., the true states are unavailable in the data~\citep{seq_cov_estimation}). 
Accordingly, we define our baseline method for this work:

\begin{method}[Estimated KF]
    \label{method:KF}
    A KF whose parameters $Q$ and $R$ were determined from data using Eq.~\ref{eq:noise_estimation}.
\end{method}

The preference of noise estimation relies on the fact that under Assumption~\ref{def:KF_assumptions} the KF -- with parameters corresponding to the noise covariances -- minimizes the square filtering errors (MSE) of the estimates of the system-states $\{X_t\}$.
Hence, equivalently, we could explicitly look for the parameters that minimize the MSE, e.g., using the Adam optimization algorithm~\citep{adam}.
Adam is a popular variant of the well-known gradient-descent algorithm, and has achieved remarkable results in the field of machine learning, including non-convex optimization problems with local-minima~\citep{adam_revisited} and millions of parameters~\citep{bert}.

Numeric optimization is more complicated than estimating two covariance matrices; however, it remains loyal to the true objective of filtering errors minimization even when the KF assumptions are violated.
Stable optimization algorithms are available in several open-source packages, e.g., PyTorch~\citep{pytorch}, which supports gradient-propagation through matrix-inversion (as needed for the KF computations).
As demonstrated in the experiments below, the PyTorch implementation of Adam provides stable results on MSE-optimization of a KF. This is conceivable, as the KF is arguably a simple model in terms of mathematical sophistication and number of parameters (e.g., in comparison to deep neural networks).

One major challenge in optimizing the KF parameters is that both $Q$ and $R$ correspond to covariance matrices, which are symmetric and positive definite (SPD): a naive numeric optimization of their entries may ruin the SPD structure.
This difficulty often motivates optimization methods that avoid gradients~\citep{Thrun_trainingKF}, or even restriction of $Q$ and $R$ to be diagonal~\citep{unsupervised_KF_tuning}.
Indeed, \citet{noise_cov_estimation} pointed out that \say{\textit{since both the covariance matrices must be constrained to be positive semi-definite, $Q$ and $R$ are often parameterized as diagonal matrices}}.

To allow Adam to optimize the non-diagonal $Q$ and $R$ we use the Cholesky parameterization~\citep{cov_parameterization}. It relies on Cholesky decomposition~\citep{cholesky}, which states that any SPD matrix $A\in \mathbb{R}^{n\times n}$ can be written as $A=LL^\top$, where $L$ is lower-triangular with positive entries along its diagonal.
The reversed claim is also true: for any lower-triangular $L$ with positive diagonal, $LL^\top$ is SPD.
Accordingly, to parameterize an SPD $A \in \mathbb{R}^{n\times n}$, we define $A(L) \coloneqq LL^\top$ and
$$\left(L(\theta)\right)_{ij}\coloneqq\begin{cases} 0 & \text{if } i<j, \\ e^{\theta_{n(n-1)/2+i}} & \text{if } i=j, \\ \theta_{(i-2)(i-1)/2+j} & \text{if } i>j, \end{cases}$$
where $\theta\in\mathbb{R}^{n(n+1)/2}$.
Clearly, the transformation $A(L(\theta)) = L(\theta)L(\theta)^\top$ is differentiable and outputs an SPD matrix for any realization of $\theta \in \mathbb{R}^{n(n+1)/2}$.
The exponent on the diagonal of $L(\theta)$ (which is an increasing and positive transformation) guarantees the uniqueness of the parameterization, though it is not necessary for the SPD property.


Note that other methods exist for SPD optimization, but require SVD-decomposition every iteration and thus are computationally heavier, e.g., matrix-exponent \citep{matrix_exponentiated_gradient_updates} and projected gradient-descent with respect to the SPD cone~\citep{projected_GD}.
The Cholesky parameterization only requires a single matrix multiplication, and thus is both efficient and easy to implement.

The Cholesky parameterization is well-known, yet to the best of our knowledge we are the first to use it for gradient-based optimization of KF.
Our approach is summarized in Method~\ref{method:OKF} and Algorithm~\ref{algo:OKF}. 

\begin{method}[Optimized KF]
    \label{method:OKF}
    A KF whose parameters $Q,R$ were determined from data of states and observations $\{(x_{k,t},z_{k,t})\}$ by solving (e.g., using Algorithm~\ref{algo:OKF}):
    $$\underset{Q',R'}{\mathrm{argmin}} \sum_{k \in \text{targets}} \sum_{t=1}^{T_k} \mathrm{loss}\left( \hat{x}_{k,t}\left(\{z_{k,\tau}\}_{\tau=1}^t;\,Q',R'\right),\ x_{k,t} \right)$$
\end{method}

\begin{algorithm}
\caption{Optimized Kalman Filter (OKF)}
\label{algo:OKF}
\DontPrintSemicolon
\SetAlgoNoLine
\SetNoFillComment

 {\bf Input}: states and observations $\{(x_{k,t},z_{k,t})\}$; dimensions $d_x,d_z$; loss function; learning rate $\alpha$\;
 \BlankLine
 \BlankLine
 Initialize $\theta_Q \in\mathbb{R}^{\frac{d_x(d_x+1)}{2}};\ \theta_R \in\mathbb{R}^{\frac{d_z(d_z+1)}{2}};\ \hat{x} \in\mathbb{R}^{d_x}$\;
 $\theta \leftarrow \text{concat}(\theta_Q,\theta_R)$\;
 \For{training epoch}{
  \For{batch in data}{
  $grad \leftarrow 0$\;
  \For{trajectory in batch}{
  \For{$t$ in trajectory}{
  $\hat{x} \leftarrow \text{KF\_predict\_and\_update}($ \newline $\hat{x}, z_{k,t};\ L(\theta_Q)L(\theta_Q)^\top,\, L(\theta_R)L(\theta_R)^\top)$\; \label{line:pred}
  $grad \leftarrow grad + \nabla_\theta \mathrm{loss}\left( \hat{x},\ x_{k,t} \right)$ \label{line:grad}\;
  }
 }
 $\theta \leftarrow \theta - \alpha \cdot grad$ \label{line:update}\;
 }
 }
 Return $Q=L(\theta_Q)L(\theta_Q)^\top,\ R=L(\theta_R)L(\theta_R)^\top$\;
\end{algorithm}

Algorithm~\ref{algo:OKF} is presented in a basic form for simplicity, but can be easily modified.
For example,
Line~\ref{line:update} can apply any update rule (e.g., in this work we use Adam update rule);
and Line~\ref{line:pred} can be separated such that the loss is calculated before the observation.
The loss itself (Line~\ref{line:grad}) is often the MSE: $\mathrm{loss}\left( \hat{x},\ x_{k,t} \right) = ||\hat{x}-x_{k,t}||^2$.


\section{Optimization vs. Noise Estimation: a Case Study}
\label{sec:OKF}

In this section, we introduce a detailed case study in the widely-researched Doppler radar problem~\citep{modern_radar,doppler2}, to compare noise estimation and errors optimization in the KF.
As mentioned in Section~\ref{sec:preliminaries}, the two are equivalent only under Assumption~\ref{def:KF_assumptions}~\citep{KF_fresh_look}.
The need to rely on many assumptions might explain why there are several extensions and design decisions in a KF configuration.
This includes the choice between KF and EKF; the choice between an educated state initialization and a simple uniform prior~\citep{KF_initialization}; and certain choices that may be made without even noticing, such as the coordinates of the state representation.
The case study below justifies the following claims:
\begin{enumerate}
    \item Design decisions in a KF are often delicate and are potentially significant.
    \item Tuning a KF by noise estimation is often sub-optimal -- even in very simple scenarios.
    \item OKF improves both accuracy and robustness to design decisions.
    \item OKF is robust to distributional shifts (Appendix~\ref{sec:generalization}) and to small training datasets (Appendix~\ref{sec:train_size}).
\end{enumerate}



\subsection{Setup and methodology}
\label{sec:setup}

\textbf{The Doppler radar problem:}
We consider a homogeneous 3D space with the radar in its origin.
Each target corresponds to a sequence of tens or hundreds states in consecutive time-steps of constant size $dt=1 sec$, and a corresponding sequence of radar measurements.
A state $x_{full}=(x_x,x_y,x_z,x_{vx},x_{vy},x_{vz})^\top\in\mathbb{R}^6$ consists of 3D location and velocity. We also denote $x=(x_x,x_y,x_z)^\top,\,u=(x_{vx},x_{vy},x_{vz})^\top$ for the separate location and velocity (or $x_{k,t}\in\mathbb{R}^3$ for the location of target $k$ at time $t$).
To simulate realistic targets, we let each target alternately perform intervals of straight motions (with or without acceleration) and turns (horizontally or vertically).
The randomization of each target determines the number of intervals, their lengths and types, the turns directions, the speed ranges and the accelerations (both in speed changes and in turns).

An observation $z\in\mathbb{R}^4$ consists of measurements of range, azimuth, elevation and Doppler signal, with an additive i.i.d Gaussian noise. Note that the former three correspond to a measurement of $x$ in spherical coordinates, and the latter one measures the projection of velocity onto the radial direction $x$.
The goal is to minimize the error of the point-estimate $\hat{x}$ of the location: $MSE=\sum_{k}\sum_t ||\hat{x}_{k,t}-x_{k,t}||^2$.

In terms of Eq.~\eqref{eq:KF_model}, the KF relies on the following model:
\begin{equation}
\label{eq:KF_for_doppler_radar}
    F=\left(\begin{smallmatrix} 1 &&& 1 && \\ & 1 &&& 1 & \\ && 1 &&& 1 \\&&& 1 &&\\&&&& 1 &\\&&&&& 1 \end{smallmatrix}\right), \ \ H=H(x)=\left(\begin{smallmatrix} 1 \\ & 1 \\ && 1 \\ &&& \frac{x_x}{r} & \frac{x_y}{r} & \frac{x_z}{r} \end{smallmatrix}\right)
\end{equation}
where $r=\sqrt{x_x^2+x_y^2+x_z^2}$. That is, the motion model is the standard constant-velocity model, and the observation model corresponds to a direct observation of $x$ and a radial projection of $u$ onto $x$.
Note that the location coordinates of $z$ are assumed to be transformed from spherical to Cartesian coordinates before multiplying them by $H$.

In contrast to the formulation of Eq.~\ref{eq:KF_model}, $H=H(x)$ depends on the unknown state $x$, hence $H(x)$ is unknown in inference time and the observation model $h(x_{full})=H(x)\cdot x_{full}$ is not linear.
To adapt the KF, we use the common approximation $H(x) \approx H(z)$ in inference, where $z$ is the recent observation ($H(\hat{x})$ was also attempted but provided inferior results).


The case study considers 5 types of tracking scenarios (\textit{benchmarks}) and 4 variants of the KF (\textit{baselines}) -- 20 experiments in total.
For each benchmark, we simulate 1500 targets for training and 1000 targets for testing.
The target state $x_{full}$ (the ground-truth) is assumed to be known on training but not on testing.
This setup is common in the non-linear filtering literature (see Section~\ref{sec:related_work}), and can be practically obtained, for example, by controlled experiments with accurate targets sensors (e.g., GPS), or by a simulation guided with domain knowledge.
The test targets differ from the training targets in the seeds used to simulate them.
In Section~\ref{sec:NKF} and Appendix~\ref{sec:generalization} we also test for generalization, where the test targets follow a different behavior than the training targets.

\textbf{Benchmarks (scenarios):}
The \textit{Free Motion} benchmark is intended to represent a realistic Doppler radar problem, with targets and observations simulated as described above.
While the KF formulation (Eq.~\eqref{eq:KF_for_doppler_radar}) appears to present quite a simple model, one may notice that multiple KF assumptions are violated in this problem (the adventurous reader may attempt to list them all before reading on). 
First, the targets follow a non-linear motion model $F$ (alternatively, if we view $F$ as linear with noise, then the noise follows the targets patterns and is not i.i.d).
Second, the observation model $H$ is not linear (as mentioned above).
Third, the distribution of the initial target state is unknown, as the standard KF implementation uses the uniform prior, whereas the simulated targets are initialized to move mostly horizontally.

In fact, the list of violations is not complete yet.
As mentioned above, the radar noise is simulated i.i.d in spherical coordinates.
However, the state is represented (and the error is measured) in Cartesian coordinates, in which the noise is \textit{not} i.i.d anymore.
To see that, consider a radar with noiseless angular estimation (i.e., only radial noise), and a low target (in the same height as the radar).
Clearly, most of the noise concentrates on the XY plane -- both in the current time-step and in the following ones (until the target moves away from the plane). Hence, the noise is statistically-dependent over time-steps.
Appendix~\ref{sec:theory_noniid} analyzes the effect of such a violation of the i.i.d assumption on the optimal KF parameters.
\textbf{This demonstrates how fragile the KF assumptions are in real-world problems -- and how their violation may sometimes be non-trivial to even notice}.

\begin{figure}
\centering
\begin{subfigure}{.49\linewidth}
  \centering
  \includegraphics[width=1.\linewidth]{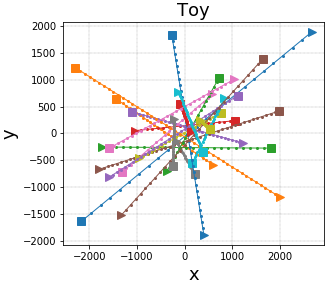}
  \caption{}
  \label{fig:trajectories_toy}
\end{subfigure}
\begin{subfigure}{.49\linewidth}
  \centering
  \includegraphics[width=1.\linewidth]{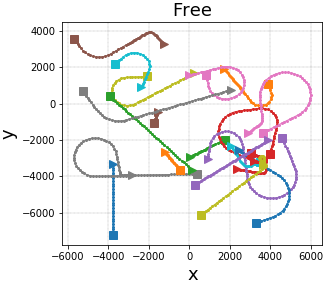}
  \caption{}
  \label{fig:trajectories_free}
\end{subfigure}
\caption{\small 
Targets in Toy and Free Motion benchmarks, projected onto XY plane.}
\label{fig:trajectories}
\end{figure}

We design 5 benchmarks to explore the effect of various assumptions violations on the KF accuracy.
Figure~\ref{fig:trajectories} displays a sample of trajectories in the simplest benchmark (\textit{Toy}), which satisfies all KF assumptions except for the non-linear $H$; and in the more realistic \textit{Free Motion} benchmark.
Samples of the other 3 benchmarks are displayed in Figure~\ref{fig:all_trajectories}. Table~\ref{tab:benchmarks} defines each benchmark more formally as a certain subset of the following properties:
\begin{itemize}
    \item \textit{anisotropic}: horizontal motion is more likely than vertical (otherwise direction is distributed uniformly).
    \vspace{-0.05cm}\item \textit{polar}: radar noise is generated i.i.d in spherical coordinates (otherwise noise is Cartesian i.i.d, in contrast to the actual physics of the system).
    \vspace{-0.05cm}\item \textit{uncentered}: targets are initialized diffusely (otherwise they are concentrated close to the radar).
    \vspace{-0.05cm}\item \textit{acceleration}: speed change is allowed (through intervals of constant acceleration).
    \vspace{-0.05cm}\item \textit{turns}: non-straight motion is allowed.
\end{itemize}
An optimal analytic solution of each benchmark is hard to derive -- even if the benchmark assumptions are noticed and modeled correctly. However, in Appendix~\ref{sec:toy_analysis} we theoretically analyze the Toy benchmark, showing that its single assumption violation is sufficient to significantly modify the optimal parameters $R$.

\begin{table}
\centering
\caption{\small Benchmarks and the properties that define them. "V" means that the benchmark satisfies the property.}
\label{tab:benchmarks}
\setlength\tabcolsep{5pt}
\begin{tabular}{|c|ccccc|}
\hline
Benchmark & \rotatebox[origin=c]{60}{anisotropic} & \rotatebox[origin=c]{60}{polar} & \rotatebox[origin=c]{60}{uncentered} & \rotatebox[origin=c]{60}{acceleration} & \rotatebox[origin=c]{60}{turns} \\
\hline
Toy & O & O & O & O & O \\
Close & V & V & O & O & O \\
Const\_v & V & V & V & O & O \\
Const\_a & V & V & V & V & O \\
Free & V & V & V & V & V \\
\hline
\end{tabular}
\end{table}

\textbf{Baselines (KF variants):}
To further demonstrate the robustness of our method, we apply it to 4 different variants of the KF.
The 4 baselines differ from each other by using either KF or EKF, with either Cartesian or spherical coordinates for representation of $R$ (the rest of the system is always represented in Cartesian coordinates).
In the EKF, the approximated observation model $H(x) \approx H(z)$ is replaced with $H(x) \approx \nabla_x h(\hat{x})$.
As mentioned above, each baseline is tuned once by Method~\ref{method:KF} and once by Method~\ref{method:OKF}.
For Method~\ref{method:OKF}, we rely on Algorithm~\ref{algo:OKF} with Adam optimizer, a single epoch, 10 targets per batch and learning rate of 0.01. 
The optimization was run for all baselines in parallel and required a few minutes per benchmark, on eight i9-10900X CPU cores in a single Ubuntu machine.

\begin{table*}
\centering
\caption{\small Summary of the errors of the various models over the various benchmarks (on out-of-sample test data). In the model names, "O" denotes optimized, "E" denotes extended, and "p" denotes polar, For example, OEKFp is an \textit{extended} KF with \textit{polar} (spherical) representation of $R$ and \textit{optimized} parameters.
For KFp, we also consider an oracle-realization of $R$ according to the true noise of the simulated radar in spherical coordinates (available only in spherical benchmarks). Evidently, for all benchmarks and all baselines, optimization yields lower errors than noise estimation.
Confidence intervals are available in Figure~\ref{fig:res_all_KF}, and are typically small in comparison to the optimization-estimation gap.
}
\label{tab:res_OKF}
\setlength\tabcolsep{5pt}
\begin{tabular}{|c|cc|ccc|cc|cc|}
\hline
Benchmark & KF & OKF & KFp & KFp (oracle) & OKFp & EKF & OEKF & EKFp & OEKFp \\
\hline
Toy & 151.7 & 84.2 & 269.6 & -- & 116.4 & 92.8 & {\bf 79.4} & 123.0 & 109.1 \\
Close & 25.0 & 24.8 & 22.6 & 22.5 & {\bf 22.5} & 26.4 & 26.1 & 24.5 & 24.1 \\
Const\_v & 90.2 & 90.0 & 102.3 & 102.3 & {\bf 89.2} & 102.5 & 99.7 & 112.7 & 102.1 \\
Const\_a & 107.5 & 101.6 & 118.4 & 118.3 & {\bf 100.3} & 110.0 & 107.0 & 126.0 & 108.7 \\
Free & 125.9 & 118.8 & 145.6 & 139.3 & {\bf 117.9} & 135.8 & 121.9 & 149.3 & 120.0 \\
\hline
\end{tabular}
\end{table*}

\subsection{Results}

\textbf{Design decisions are not trivial:}
Table~\ref{tab:res_OKF} summarizes the tracking errors. The left column in each cell corresponds to Method~\ref{method:KF} (standard KF), and shows that in each benchmark, the errors strongly depend on the design decisions ($R$'s coordinates and whether to use EKF).
In the Toy benchmark, for example, EKF is the best design, since the observation model $H$ is non-linear.
In other benchmarks, however, the winning designs of the non-optimized KF are arguably surprising:
\begin{enumerate}
    \item Under non-isotropic motion direction (all benchmarks except Toy), EKF is worse than KF despite the non-linearity. It is possible that the horizontal prior reduces the stochasticity of $H$, making the derivative-based approximation unstable.
    \item Even when the observation noise is spherical i.i.d, spherical representation of $R$ is not beneficial when targets are scattered far from the radar (last 3 benchmarks in Table~\ref{tab:res_OKF}). It is possible that with distant targets, Cartesian coordinates have a more important role in expressing the horizontal prior of the motion.
\end{enumerate}
Since the best KF variant per benchmark seems hard to predict in advance, a practical system cannot rely on choosing the KF variant optimally -- and should rather be robust to this choice.

\textbf{Optimization is more accurate and robust:}
Table~\ref{tab:res_OKF} shows that for \textit{every} benchmark and \textit{every} baseline (20 experiments in total), OKF yielded smaller errors than noise estimation (over an out-of-sample test dataset).
Note that OKF wins even in the Toy scenario, under the slightest violation of KF assumptions. 
In addition, the variance between the baselines reduces under optimization, i.e., OKF makes the KF more robust to design decisions. 

\textbf{Oracle noise estimation baseline:}
We studied the performance of a KF with a perfect knowledge of the noise covariance matrix $R$. Note that in the constant-speed benchmarks, the estimation of $Q=0$ is already very accurate, hence in these benchmarks the oracle-KF has a practically perfect knowledge of both noise covariances.
Nonetheless, Table~\ref{tab:res_OKF} shows that the oracle yields very similar results to a KF with estimated parameters.
Indeed, \textbf{the benefit of OKF is not in estimating $Q$ and $R$ more accurately, but rather in optimizing the desired objective}.


\begin{figure}[!h]
\centering
\begin{subfigure}{.49\linewidth}
  \centering
  \includegraphics[width=1.\linewidth]{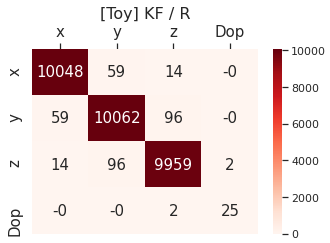}
  \caption{\small Estimated $R$}
  \label{fig:noise_R_KF_toy}
\end{subfigure}
\begin{subfigure}{.49\linewidth}
  \centering
  \includegraphics[width=1.\linewidth]{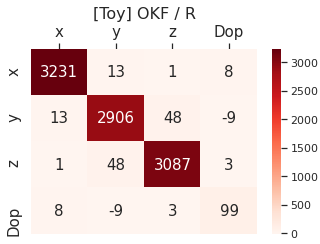}
  \caption{\small Optimized $R$}
  \label{fig:noise_R_OKF_toy}
\end{subfigure}
\caption{\small The covariance matrix $R$ of the observation noise obtained in a (Cartesian) KF by noise estimation and by optimization, based on the dataset of the Toy benchmark.
The axes correspond to the observation variables associated with the matrix entries.
Note that the noise estimation is quite accurate, as the true variance of the noise is $100^2$ for the positional dimensions and $5^2$ for Doppler.
The optimization increases the variance associated with the Doppler signal, as predicted by Lemma~\ref{lemma:toy_effective_noise}. The decrease in the other diagonal components is discussed in Appendix~\ref{sec:toy_analysis}.}
\label{fig:noise_params_toy}
\end{figure}

\textbf{Toy scenario: why is noise estimation sub-optimal?}
The gap between estimated and optimized noise parameters can be studied through the simplest Toy benchmark, where the only KF assumption violation is the non-linear $H$.
Since the unknown entries of $H$ correspond to the Doppler observation, the non-linearity inserts uncertainty to the \textit{transformation} of the Doppler observation (in addition to the noise in the observation itself). This increases Doppler's \textit{effective} noise in comparison to the location observation, as shown theoretically by Lemma~\ref{lemma:toy_effective_noise} in the appendix.
This explanation is consistent with Figure~\ref{fig:noise_params_toy}: the noise associated with Doppler is indeed increased by the optimization.
Note that the non-linearity modifies the effective noise in a delicate way, which would \textit{not} be compensated by a naive trial and error of noise inflation or deflation.

In Section~\ref{sec:more_experiments}, OKF is also shown to reduce the MSE for video tracking (MSE reduction of 18\%) and lidar-based state estimation (15\%).
In the appendix, it is further demonstrated to be robust to major {distributional shifts} and to {small training datasets}.

\section{Neural KF: Is Non-linearity Helpful?}
\label{sec:NKF}

\begin{figure}
\centering
\includegraphics[width=\linewidth]{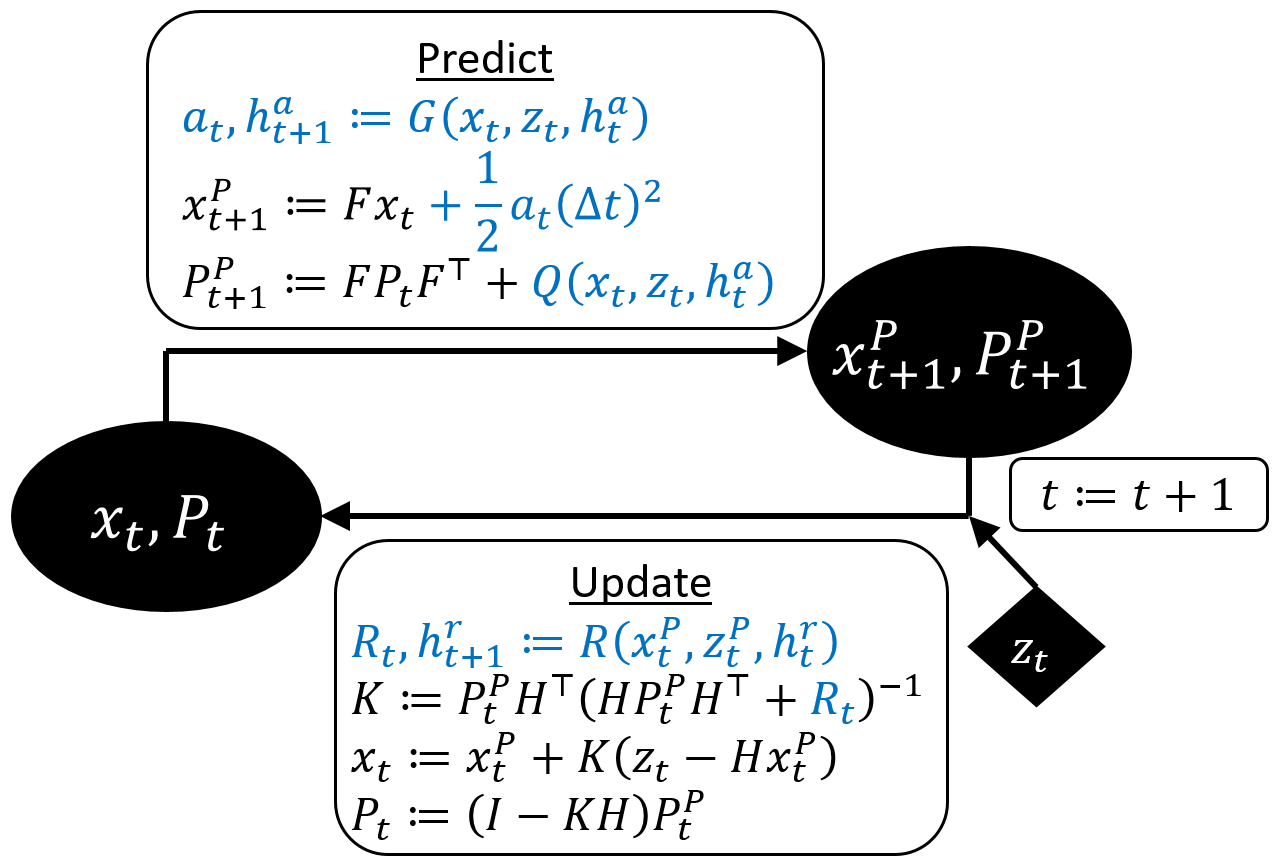}
\caption{\small The Neural Kalman Filter (NKF) algorithm. The differences from Figure~\ref{fig:KF} are marked. $\Delta t$ is a constant, $G,Q,R$ are LSTM networks, and $h^a,h^r$ are their hidden states (more accurately, $G$ and $Q$ are different heads of the same network, and thus share the same hidden states $h^a$). In addition to the raw $x_t,z_t$, the networks are also fed with certain manually-crafted features (e.g., the projection of $z_t-x_t$ on the perpendicular direction $x_t^\perp$).}
\label{fig:NKF_diagram}
\end{figure}

LSTM is an architecture of recurrent neural networks, and is a key component in many SOTA algorithms for non-linear sequential prediction~\citep{process_prediction_review} (see more details in Appendix~\ref{sec:preliminaries_extended}).
In its basic form, unlike the KF, it can only provide a point estimate (rather than predicting the whole distribution); and does not provide separate estimates for prediction (before observing a new input) and filtering (after observation).
In this section, we introduce the {\bf \textit{Neural Kalman Filter}} model (NKF), which keeps these benefits of the KF framework, while relying on an LSTM to model non-linear motion.



Every prediction step, NKF uses a neural network model to predict the target's acceleration $a_t$ and the motion uncertainty $Q_t$. Every update step, another network predicts the observation uncertainty $R_t$ (a neural correction of $H$ was also attempted but yielded inferior unstable predictions, as demonstrated in Appendix~\ref{sec:NKF_extended}).
The predicted quantities are incorporated into the standard update equations of the KF, as shown in Figure~\ref{fig:NKF_diagram}.
Note that the neural network predicts the acceleration rather than directly predicting the location. This is intended to regularize the predictive model and to exploit the domain knowledge of kinematics.

\begin{figure}[!h]
\centering
\begin{subfigure}{\linewidth}
  \centering
  \includegraphics[width=.75\linewidth]{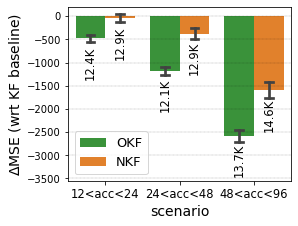}
  \caption{}
  \label{fig:res_NKF_MSE}
\end{subfigure} \\
\begin{subfigure}{\linewidth}
  \centering
  \includegraphics[width=0.9\linewidth]{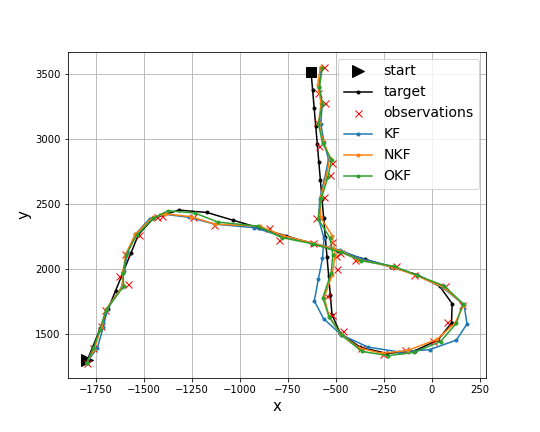}
  \caption{}
  \label{fig:res_sample}
\end{subfigure}
\caption{\small (a) Relative tracking errors (lower is better) with relation to a standard KF, over targets with different ranges of acceleration. The error-bars represent confidence intervals of 95\%. The label of each bar represents the corresponding \textit{absolute} MSE ($\times10^3$).
In the training data, the acceleration was limited to 24-48, hence the other ranges measure generalization. While the Neural KF (NKF) is significantly better than the standard KF, its advantage is entirely eliminated once we optimize the KF (OKF). (b) A sample target and the corresponding models outputs (projected onto XY plane). The standard KF has a difficulty to handle some of the turns.}
\label{fig:res_NKF}
\end{figure}

After training NKF on a dataset of simulated targets with random turns and accelerations (similarly to the Free Motion benchmark of Section~\ref{sec:OKF}), we tested it on a test set, where the targets have different seeds and an extended range of permitted accelerations (both speed-changes and turns-sharpness).
As shown in Figure~\ref{fig:res_NKF}, NKF significantly reduces the tracking errors compared to a standard KF.


At this point, it seems that the non-linear architecture of NKF provides better accuracy in the non-linear problem of radar tracking.
However, Figure~\ref{fig:res_NKF} shows that when comparing NKF to an \textit{optimized} KF (OKF), we \textit{completely eliminate} the advantage of NKF, and in fact reduce the errors even further.
In other words, in this experiment the benefits of NKF come \textit{only} from optimization and \textit{not at all} from the expressive architecture.
By overlooking the sub-optimality of noise estimation in the KF, we would wrongly adopt the over-complicated NKF.

Note that OKF also generalizes well to targets with different accelerations than observed in the training, which indicates a certain robustness to distributional shifts.
In Appendix~\ref{sec:NKF_extended}, we present extended experiments with different variants of NKF, another tracking benchmark, and a likelihood evaluation metric (NLL) in addition to estimation error (MSE). Note that high likelihood score can improve the matching in a multi-target tracking problem.

Of course, our results do not imply that neural-networks in general cannot be superior to a KF: only that when comparing the two, if the KF is not optimized similarly to the neural model, the experimental results may be misleading.
As discussed in Section~\ref{sec:related_work}, this wrong \textit{methodology} is not uncommon in the literature.


\section{OKF in Other Domains}
\label{sec:more_experiments}

Section~\ref{sec:OKF} demonstrates the sub-optimality of noise estimation in KF, as well as the benefits of OKF, in the Doppler radar tracking problem.
This section extends the experiments to 2 additional domains, corresponding to different violations of the KF assumptions. In particular, in contrast to the Doppler radar, both domains correspond to a linear observation model.

\subsection{Video Tracking}
\label{sec:MOT20}


For the benchmark of tracking targets in a video we consider the MOT20 dataset~\citep{MOT20}.
MOT20 includes several videos with multiple targets to track (mostly pedestrians).
The dataset also includes the ground-truth location and size of the targets in every frame of the videos.
As object detection is out of our scope, we consider these ground-truth states as direct observations, i.e., we assume to have a detector with zero observation error (which in particular corresponds to a linear observation model, in contrast to Section~\ref{sec:OKF}).

\begin{figure}[!h]
\centering
\includegraphics[width=1\linewidth]{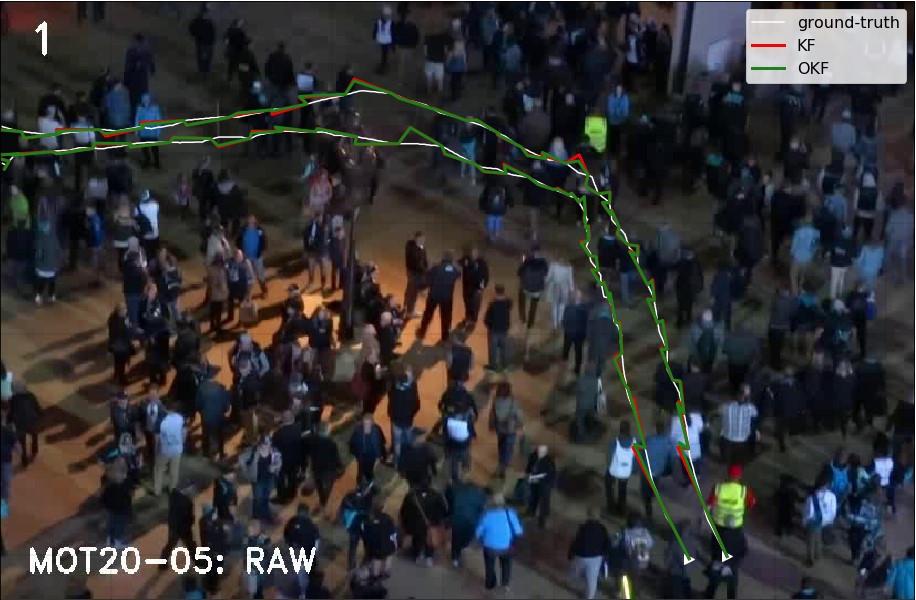}
\caption{\small A sample of 2 targets in the first frame of MOT20 test video. The true targets trajectories are shown along with the predictions of KF and OKF (each prediction is done one time-step in advance).}
\label{fig:MOT20_sample}
\end{figure}

More specifically, we characterize a target using the state $x=(x_x,x_y,x_w,x_h,x_{vx},x_{vy})\in \mathbb{R}^6$ (2D location, size and velocity), and an observation as $z=(z_x,z_y,z_w,z_h)$. Note that location and size are provided in the dataset, and velocity is derived accordingly. The observation model is described above, and for the motion model we assume constant velocity and constant target size, leading to:
$$ F=\left(\begin{smallmatrix} 1&&&&1& \\ &1&&&&1 \\ &&1&&& \\ &&&1&& \\ &&&&1& \\ &&&&&1 \end{smallmatrix}\right), \quad H=\left(\begin{smallmatrix} 1&&&&0&0 \\ &1&&&0&0 \\ &&1&&0&0 \\ &&&1&0&0 \end{smallmatrix}\right) $$

For the train data we use the videos MOT20-01,MOT20-02,MOT20-03, and for the test data MOT20-05. In particular, our test data comes from an entirely different video than the train data, hence the testing is less prone to overfit.
The optimization objective is the MSE of the location $(x_x,x_y)$ after the prediction step (note that in absence of observation noise, the error of the update step is simply zero).

\begin{figure}
  \begin{center}
  \includegraphics[width=0.7\linewidth]{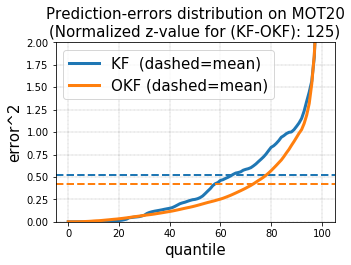}
  \end{center}
\caption{\small Prediction errors of KF and OKF on 1208 targets of the test data of MOT20 videos dataset. The MSE of OKF is smaller by 18\%, with statistical significance of $\text{p-value}<10^{-6}$.}
\label{fig:MOT20_res}
\end{figure}

As shown in Figure~\ref{fig:MOT20_res}, OKF reduces the prediction MSE by 18\%.
The significance level of the results is also indicated by the z-value $z=125$, corresponding to the errors difference (calculated as $z=mean(\Delta err)/std(\Delta err)\cdot\sqrt{N_{samples}}$).




\subsection{Lidar-based State Estimation in Self Driving}
\label{sec:lidar}

In this section, we test OKF in a benchmark of state-estimation in self-driving, based on lidar measurements with respect to known landmarks~\citep{lidar_from_beacons}.
For simplicity, we focus on location estimation and on a single landmark (as the matching problem from multiple landmarks is out of the scope of this work).

\begin{figure}[!h]
\centering
\begin{subfigure}{.45\linewidth}
    \includegraphics[width=\linewidth]{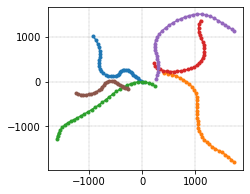}
    \caption{}
    \label{fig:lidar_trajectories}
\end{subfigure} \\
\begin{subfigure}{.53\linewidth}
    \includegraphics[width=\linewidth]{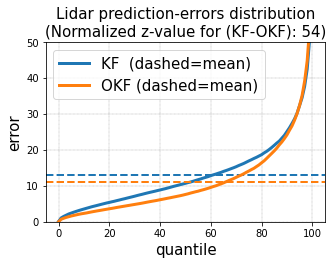}
    \caption{}
    \label{fig:lidar_MSE}
\end{subfigure}
\begin{subfigure}{.45\linewidth}
    \includegraphics[width=\linewidth]{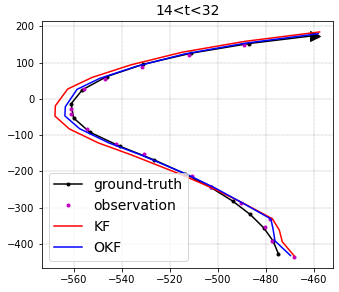}
    \caption{}
    \label{fig:lidar_sample}
\end{subfigure}
\caption{\small (a) A sample of 6 targets trajectories from the simulated self-driving data. (b) Prediction errors of KF and OKF on 600 targets of the test data. The statistical significance of OKF's improved accuracy is indicated by the z-value of the errors difference (calculated as $z=mean(\Delta err)/std(\Delta err)\cdot\sqrt{N}$). (c) the tracking of KF and OKF on a sample target trajectory, focused on the time interval $14<t<32$ for visual clarity.}
\end{figure}

We formulate the problem as follows: a vehicle state is $x=(x_x,x_y,x_{vx},x_{vy})\in \mathbb{R}^4$ (2D location and velocity), and an observation is $z=(z_x,z_y)$. The dynamics are simulated as follows: each target trajectory consists of several intervals, in each one the target has certain (positive or negative) acceleration and certain lateral acceleration ("turn magnitude"), both drawn randomly at the beginning of the interval. The observation noise is drawn i.i.d in polar coordinates.
We simulate a corresponding dataset of 2000 targets as demonstrated in Figure~\ref{fig:lidar_trajectories}, and split them to train set (1400 targets) and test set (600 targets).

From the KF point-of-view, the dynamics and observations are modeled by:
$$ F=\left(\begin{smallmatrix} 1&0&1&0 \\ 0&1&0&1 \\ 0&0&1&0 \\ 0&0&0&1 \end{smallmatrix}\right), \quad H=\left(\begin{smallmatrix} 1&0&0&0 \\ 0&1&0&0 \end{smallmatrix}\right) $$
Note that in contrast to the Doppler radar, this benchmark has a linear observation model; and in contrast to the video tracking, the observation is not noiseless.
On the other hand, similarly to the Doppler radar, the Cartesian representation of the observations eliminates their i.i.d property.

We trained OKF with respect to the MSE of the estimated target location.
The optimization significantly reduced the errors by 14.9\%, as shown in Figure~\ref{fig:lidar_MSE}.
In addition, Appendix~\ref{sec:theory_noniid} provides theoretical analysis for the optimal parameters under the coordinates mismatch between the state and the source of the noise.


\section{Related Work}
\label{sec:related_work}

\textbf{Noise estimation:}
When tuning a KF, ground-truth states data is often unavailable~\citep{noise_cov_estimation}.
Thus, estimation of the KF noise parameters from observations alone has been studied for decades, addressed using various methods based on autocorrelation~\citep{Mehra,Carew}, EM~\citep{EM_for_noise_estimation} and others~\citep{ALS,seq_cov_estimation,measurement_noise_recommendation}.
However, if the states \textit{are} available, noise estimation reduces to Eq.~\ref{eq:noise_estimation} and is considered a solved problem~\citep{ALS}. In this case, we show that although noise estimation is indeed easy, it is often not the right task to approach.

Many works addressed the problem of non-stationary noise estimation~\citep{cov_estimation_varying_processes,adaptive_noise_covariance}.
However, as demonstrated in Section~\ref{sec:NKF}, in certain cases stationary methods are highly competitive if tuned correctly -- even in problems with complicated dynamics.

\textbf{Optimization:}
In this work we apply gradient-based optimization to the KF with respect to its errors.
Optimization without gradients was already suggested in \citet{Thrun_trainingKF}.
In practice, "optimization" of the KF is often handled manually using trial-and-error~\citep{learning_in_indoor_navigation} or a grid-search over possible values of $Q$ and $R$~\citep{noise_cov_estimation,pose_estimation}.
In other cases, $Q$ and $R$ are restricted to be diagonal~\citep{unsupervised_KF_tuning,noise_cov_estimation}.

Gradient-based optimization of SPD matrices in general was suggested in \citet{matrix_exponentiated_gradient_updates} using matrix-exponents, and is also possible using projected gradient-descent~\citep{projected_GD} -- both rely on SVD-decomposition.
In this work, we apply gradient-based optimization using the parameterization that was suggested in \citet{cov_parameterization}, which requires a mere matrix multiplication, and thus is both efficient and easy to implement.

In absence of a trajectories dataset, a recent line of works~\citep{tsiamis2020online, goel2021regret} minimizes a regret metric, using online optimization from the observations of the current trajectory.

\textbf{Neural Networks (NNs) in filtering problems:}
Section~\ref{sec:NKF} presents a RNN-based extension of the KF, and demonstrates how its advantage over the linear KF vanishes once the KF is optimized.
The use of NNs for non-linear filtering problems is very common in the literature, e.g., in online tracking prediction~\citep{ManeuveringTargetTracking2,KF_vs_LSTM,pose_estimation,navigation_using_RNN,KF_with_ANN}, near-online prediction~\citep{vision_tracking_with_bilinear_LSTM}, and offline prediction~\citep{DeepMTT}.
In addition, while \citet{sort} apply a KF for video tracking from mere object detections, \citet{deep_sort} add to the same system a NN that generates visual features as well.
NNs were also considered for related problems such as data association~\citep{DeepDA}, model-switching~\citep{IMM_with_RNN_transitions}, and sensors fusion~\citep{sensor_fusion}.

Many works that consider NNs for filtering problems, use a KF as a baseline for comparison.
However, while the NN parameters are typically optimized with respect to the prediction errors, the KF parameters tuning is sometimes ignored \citep{ManeuveringTargetTracking2,bai2020neuron,RNN_EKF}, sometimes based on estimation (or knowledge) of the noise \citep{navigation_using_RNN,ANN_vs_EKF,KalmanNet}, and sometimes optimized heuristically as mentioned above \citep{learning_in_indoor_navigation,pose_estimation,KF_with_ANN}.
Our findings imply that this methodology is wrong, since the baseline is not optimized to the same level as the learning model.
\citet{KF_vs_NN_for_batteries} explicitly discusses the sensitivity of EKF to the noise model accuracy, and suggests the solution of a NN with supervised learning -- without ever considering the same supervised learning for the EKF itself.


\section{Conclusion}
\label{sec:summary}

Through a detailed case study, we demonstrated both theoretically and empirically the fragility of the KF assumptions, and how the slightest violation of them may change the effective noise in the problem -- leading to significant and non-trivial changes in the optimal noise parameters.
OKF addresses this problem using optimization tools from supervised machine learning, and uses Cholesky parameterization to apply them efficiently to the SPD parameters of the KF.

We demonstrated the accuracy of OKF over different violations of KF assumptions, in different domains (radar tracking, video tracking and lidar-based state estimation), with relation to different KF variants, over small and large training datasets, and even under distributional shifts between train and test datasets.
Indeed, once we acknowledged the need for optimization, OKF provided it robustly over all these scenarios.


We also discussed one of the consequences of using a sub-optimal KF: the common methodology of comparing an optimized filtering algorithm to a classic variant of the KF is misleading, unless the KF is optimized in a similar manner.
We demonstrated that once the KF is optimized, it may become competitive with a non-linear model such as LSTM -- even in a non-linear problem.

In light of this evidence, we recommend to use OKF as the default procedure for KF tuning in presence of ground-truth data, whenever the KF assumptions are not strictly-guaranteed.
To provide a practical solution in addition to the conceptual algorithm, we publish an off-the-shelf implementation as a free \href{https://pypi.org/project/Optimized-Kalman-Filter/}{\underline{PyPI}} package.





%



\ifCLASSOPTIONcompsoc
  \section*{Acknowledgments}
\else
  \section*{Acknowledgment}
\fi

The authors thank Tal Malinovich, Mark Kozdoba, Ophir Nabati and Zahar Chikishev for their helpful advice.


\ifCLASSOPTIONcaptionsoff
  \newpage
\fi



\bibliographystyle{IEEEtran}



\bibliography{main}


%

\begin{IEEEbiography}[{\includegraphics[width=1in]{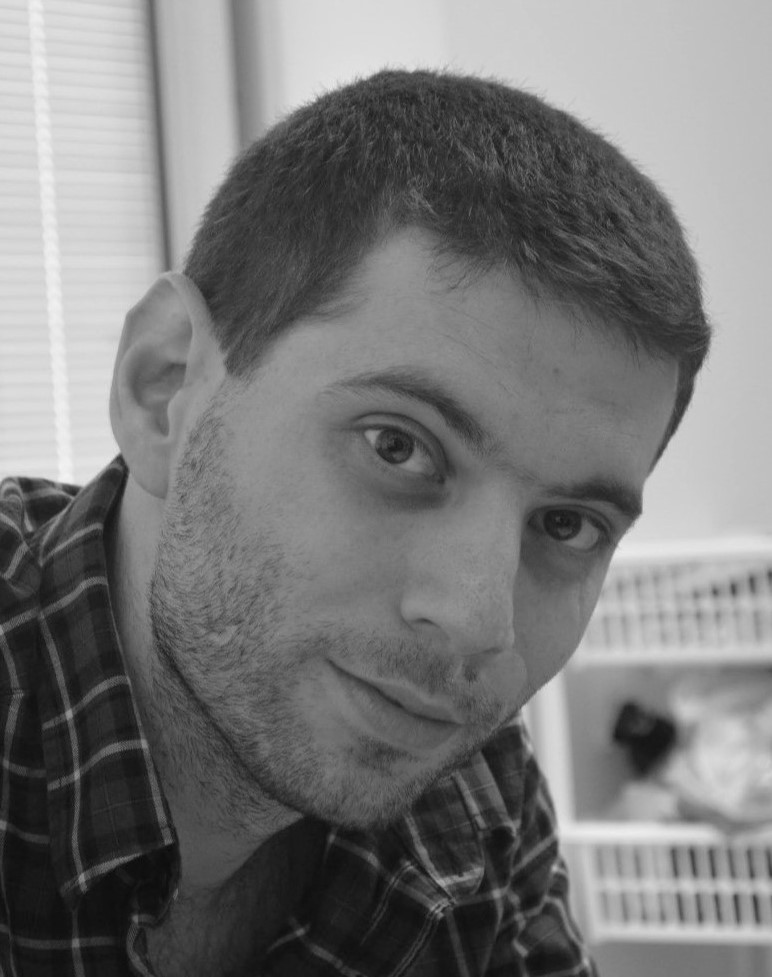}}]{Ido Greenberg}
received the B.Sc.~degree in mathematics and physics from the Hebrew University
of Jerusalem, Israel, in 2011, and the M.Sc.~degree in applied mathematics from Tel Aviv University, Israel, in 2017.
He is currently working toward the Ph.D.~degree in electrical and computer engineering at the Technion Institute of Technology, Israel.
His research interests include reinforcement learning, deep learning and machine intelligence.
\end{IEEEbiography}

\begin{IEEEbiography}[{\includegraphics[width=1in,clip,keepaspectratio]{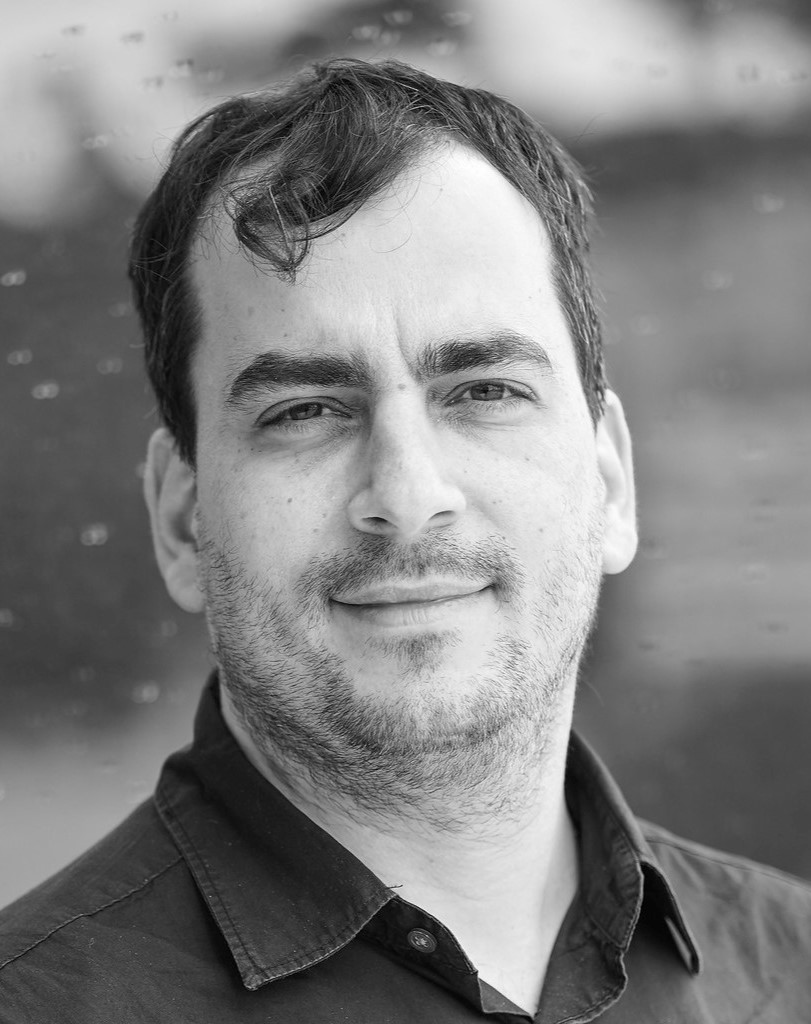}}]{Shie Mannor} (FIEEE)
received the B.Sc.~degree in electrical engineering, the B.A.~degree in mathematics, and the Ph.D.~degree in electrical engineering from the Technion—Israel Institute of Technology, Haifa, Israel, in 1996, 1996,
and 2002, respectively. From 2002 to 2004, he was
a Fulbright Scholar and a Post-Doctoral Associate
with MIT. From 2004 to 2010, he was with the
Department of Electrical and Computer Engineering, McGill University, where he was the Canada
Research Chair in machine learning. Since 2008,
he has been with the Faculty of Electrical Engineering, Technion—Israel
Institute of Technology, where he is currently a Professor.
His research interests include machine learning and pattern recognition,
planning and control, multi-agent systems, and communications.
\end{IEEEbiography}


\begin{IEEEbiography}[{\includegraphics[width=1in]{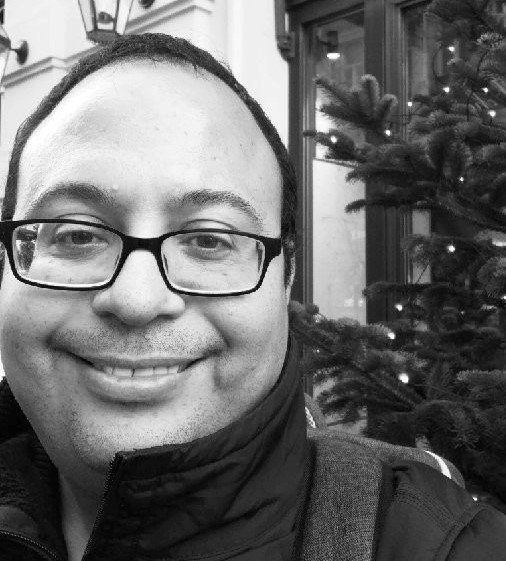}}]{Netanel Yannay}
received the B.Sc.~degree in computer engineering from the Holon Institute of Technology, Israel, in 2006.
From 2006, he has been an engineer, a team leader and a researcher in ELTA Systems Ltd.
Since 2021, he is the head of the AI research group in ELTA Systems Ltd.
His research interests include signal processing, time series forecasting and machine learning.
\end{IEEEbiography}





\setcounter{tocdepth}{1}
\tableofcontents

\newpage
\appendices
\onecolumn


\section{Theoretical Analysis: Non-linear Observation}
\label{sec:toy_analysis}

Section~\ref{sec:OKF} demonstrates that the optimal parameters in KF often differ from the covariance of the noise.
Here we analyze this effect for the private case of the Toy benchmark; that is, we prove that even in the toy problem, a single assumption violation (non-linear observation model $H(X)$) may significantly modify the optimal parameters.

\begin{definition}[The toy Doppler problem]
\label{def:toy_problem}
    The toy Doppler problem is the filtering problem modeled by Eq.~\ref{eq:KF_model} with $F,H$ defined by Eq.~\eqref{eq:KF_for_doppler_radar} and $Q,R$ defined by
    $$ Q=\pmb{0}\in\mathbb{R}^{6\times6} \qquad R=\left(\begin{smallmatrix} \sigma_x^2 \\ & \sigma_y^2 \\ && \sigma_z^2 \\ &&& \sigma_D^2 \end{smallmatrix}\right) $$
    where $\sigma_x,\sigma_y,\sigma_z,\sigma_D>0$.
\end{definition}


Since $H(x)$ of Eq.~\eqref{eq:KF_for_doppler_radar} is unknown on inference, Section~\ref{sec:setup} uses the approximation $H(x) \approx H(z)$. Here we use a more general approximation $H(x) \approx \tilde{H}$, where
$$ \tilde{H} = \begin{pmatrix} 1 \\ & 1 \\ && 1 \\ &&& \tilde{x}_x/\tilde{r} & \tilde{x}_y/\tilde{r} & \tilde{x}_z/\tilde{r} \end{pmatrix} $$
and $\tilde{x}$ is any estimator of $x$ (e.g., the observation $z$ or the KF estimator $\hat{x}$).

\begin{lemma}[Effective observation noise in the toy Doppler problem]
    \label{lemma:toy_effective_noise}
    Consider the problem of Definition~\ref{def:toy_problem} with the approximated observation model $\tilde{H}$.
    At a certain time-step, assume that the positional estimation errors $\tilde{x}-x$ are independent of the target velocity $u$.
    Then, the observation noise corresponding to $\tilde{H}$ (the \textit{effective} observation noise) is described by the following covariance matrix:
    \begin{equation}
    \label{eq:R}
        \tilde{R} = \left(\begin{smallmatrix} \sigma_x^2 \\ & \sigma_y^2 \\ && \sigma_z^2 \\ &&& \sigma_D^2 + C \end{smallmatrix}\right) = R + \left(\begin{smallmatrix} 0 \\ & 0 \\ && 0 \\ &&& C \end{smallmatrix}\right)
    \end{equation}
    for some $C>0$.
\end{lemma}
\begin{proof}
    Denoting the normalized estimation error $dx'=\frac{\tilde{x}}{\tilde{r}}-\frac{x}{r}$, we can rewrite $\tilde{H}$ as
    \begin{equation*}
        \tilde{H} = H + 
        \left(\begin{smallmatrix}
        0 \\ & 0 \\ && 0 \\ &&& dx_x' & dx_y' & dx_z'
        \end{smallmatrix}\right)
    \end{equation*}
    By shifting the observation model in Eq.~\ref{eq:KF_model} from $H$ to $\tilde{H}$, and denoting the noise by $\nu=(\nu_x,\nu_y,\nu_z,\nu_D)^\top$, we receive
    \begin{align*}
        Z=&H\cdot x +\nu = \tilde{H}X + \left(\begin{smallmatrix} \nu_x \\ \nu_y \\ \nu_z \\ \nu_D -dx_x'u_x-dx_y'u_y-dx_z'u_z \end{smallmatrix}\right) = \tilde{H}X + \left(\begin{smallmatrix} \nu_x \\ \nu_y \\ \nu_z \\ \nu_D - dx'\cdot u \end{smallmatrix}\right)
    \end{align*}
    that is, the effective observation noise with respect to $\tilde{H}$ is $\tilde{\nu} = (\nu_x, \nu_y, \nu_z, \nu_D - dx'\cdot u)^\top$.
    
    $dx_x',\nu_x$ are both independent of the velocity $u_x$ (the former by the lemma's assumption, and the latter by the model of Eq.~\ref{eq:KF_model} in Definition~\ref{def:toy_problem}). Thus, $Corr(dx_x'\cdot u_x, \nu_x)=E(dx_x'\cdot u_x \cdot \nu_x)=E(dx_x'\nu_x)E(u_x)=0$, and similarly for $y,z$.
    Hence, by denoting $C=Var(dx'\cdot u)>0$ we have $Cov(\tilde{\nu}) = \tilde{R}$.
\end{proof}

\textbf{Empirical results:}
Lemma~\ref{lemma:toy_effective_noise} provides the theoretical explanation for Figure~\ref{fig:noise_R_OKF_toy}, where we see that the optimization of $R$ increases the variance of the Doppler signal compared to the variance of the positional signal.
Note that Figure~\ref{fig:noise_R_OKF_toy} shows not only an increased Doppler variance, but also a decreased positional variance, which is not explained by the analysis above. This is caused by the fact that when $Q\equiv0$, the absolute values of $R$ have a negligible effect compared to the relative values between the components -- hence the optimization increases Doppler's variance compared to positional variance, but is indifferent to their scale. Indeed, we tried to re-scale $R$ accordingly, and received indistinguishable empirical results.

\textbf{Effective-noise estimation:}
In the Toy benchmark, the discrepancy between optimization and noise estimation could be eliminated using a corrected noise estimation calculated with respect to $\tilde{H}$.
This is indeed a toy problem, and once modeled correctly, it can be solved analytically.
This is not the case for most of the 20 different experiments of Section~\ref{sec:OKF}.
\textbf{By optimizing from data, we bypass both modeling misspecification and analysis difficulties}.



\section{Theoretical Analysis: Non-i.i.d Noise}
\label{sec:theory_noniid}

The KF model assumes i.i.d noise in motion and observation.
Other simple noise models can sometimes be solved too -- if modeled correctly.
For example, if the noise is auto-regressive with a known order $p$, an adjusted KF model may consider the last $p$ values of the noise itself as part of the system state~\citep{KF_colored_noise}.
However, the actual noise model is often unknown or hard to solve analytically.
In Section~\ref{sec:setup} we even discuss the possibility of being entirely unaware to the violation of the i.i.d assumption.
In any of these cases, it is natural to use the standard KF model.
However, the noise covariance matrices are not guaranteed anymore to be the optimal values of the KF parameters $Q$ and $R$.

In this section, we analyze the shift in the optimal parameters following a violation of the i.i.d assumption.
We consider the problem of a moving target in a 2D space, with observation noise drawn i.i.d in polar coordinates (similarly to the lidar model of Section~\ref{sec:lidar}).
For simplicity of the analysis we represent the state using only the location $x=(x_1,x_2)^\top\in\mathbb{R}^2$, and assume a no-motion model up to an isotropic noise. Furthermore, we assume that the observation noise has only a radial component.

\begin{definition}[The simplified lidar problem]
\label{def:simplified_lidar}
    The simplified lidar problem is the filtering problem modeled by Eq.~\ref{eq:KF_model} with the following parameters:
    $$F=H=\begin{pmatrix} 1 & 0 \\ 0 & 1 \end{pmatrix} \quad Q=\begin{pmatrix} q & 0 \\ 0 & q \end{pmatrix} \quad R_{polar}=\begin{pmatrix} r_0 & 0 \\ 0 & 0 \end{pmatrix}$$
    for some $q,r_0>0$, where the observation noise $\nu_t$ is drawn i.i.d from $N(0,R_{polar})$ in \textit{polar} coordinates.
    We also assume that the initial system-state $x_0$ is drawn from an isotropic distribution, i.e., follows a density function of the form $f((x_0)_1^2+(x_0)_2^2)$, which is invariant to the target direction.
\end{definition}

At a first glance, this simplistic problem may seem to satisfy all the KF assumptions. In particular, both sources of noise are i.i.d. However, the observation noise is drawn i.i.d in \textit{polar} coordinates, while the state is estimated (and the error is measured) in \textit{Cartesian} coordinates. As discussed in Section~\ref{sec:setup}, the i.i.d property is not invariant to the coordinates transformation!

\begin{lemma}
    The observation noise in the simplified lidar problem is not i.i.d in Cartesian coordinates.
\end{lemma}
\begin{proof}
    Denote the system state at time $t$ by $x_t=((x_t)_1,(x_t)_2)^\top$, and denote $\tan\theta=((x_t)_2/(x_t)_1)$.
    From Definition~\ref{def:simplified_lidar}, by direct coordinates transformation, the observation noise is drawn from the distribution $\nu_t\sim N(0,R(\theta))$, where
    $$R(\theta)=\begin{pmatrix} r_0\cos^2(\theta) & r_0\cos(\theta)\sin(\theta) \\ r_0\cos(\theta)\sin(\theta) & r_0\sin^2(\theta) \end{pmatrix}$$
    
    This noise covariance matrix varies every time-step with $\theta$, hence the noise $\nu_t$ is not identically-distributed in Cartesian coordinates.
    In addition, consecutive steps are likely to have similar values of $\theta$, hence the noise is not independent either.
\end{proof}

If we are aware of the particular noise model of the problem, we can estimate $R_{polar}$ in polar coordinates and transform it to the Cartesian $R(\theta)$ dynamically every time-step (e.g., using our the current estimate of $\theta$, since the true $\theta$ is unknown). This is indeed what we do in the KFp model in Section~\ref{sec:OKF}.
If we are not aware, however, we would simply estimate a constant covariance matrix $R$ in Cartesian coordinates.

\begin{lemma}[Noise estimation in the simplified lidar problem]
    \label{lemma:lidar_KF}
    Let $R$ be the sample covariance matrix of the observation noise, calculated by Eq.~\ref{eq:noise_estimation} with respect to a dataset of i.i.d targets in the simplified lidar problem.
    Then, as the number of targets in the data grows, $R$ converges almost surely to
    $$ R_{est} = \begin{pmatrix} r_0/2 & 0 \\ 0 & r_0/2 \end{pmatrix} $$
\end{lemma}
\begin{proof}
    From Definition~\ref{def:simplified_lidar}, the distribution of the initial system-states is isotropic, and so is the motion model $Q$. Hence, for any target at any time-step, the distribution of its direction $\theta$ (defined by $\tan\theta=((x_t)_2/(x_t)_1)$) is uniform: $\theta\sim\left[0,2\pi\right)$.
    Now we can calculate:
    \begin{align*}
        E_\theta\left[(R_{est})_{11}\right] &= E_\theta\left[r_0 \cos^2\theta\right] = \int_0^{2\pi} \frac{r_0}{2\pi} \cos^2\theta d\theta = \frac{r_0}{2} \\
        E_\theta\left[(R_{est})_{22}\right] &= E_\theta\left[r_0 \sin^2\theta\right] = \int_0^{2\pi} \frac{r_0}{2\pi} \sin^2\theta d\theta = \frac{r_0}{2} \\
        E_\theta\left[(R_{est})_{12}\right] &= E_\theta\left[(R_{est})_{21}\right] = E_\theta\left[r_0 \cos\theta\sin\theta\right] = 0 
    \end{align*}
    and since the targets are i.i.d, by the Law of Large Numbers we have
    $$ R \xrightarrow{\text{a.s.}} R_{est} = \begin{pmatrix} r_0/2 & 0 \\ 0 & r_0/2 \end{pmatrix} $$
\end{proof}

So far we saw that the actual observation noise is $R(\theta)$ (where $\theta$ changes every time-step), while the constant covariance matrix that would be obtained by noise estimation is (asymptotically) $R_{est}$.
Assuming that we constraint ourselves to a constant matrix (invariant to $\theta$) in Cartesian coordinates, can we still do better than $R_{est}$?
More specifically, due to the radial symmetry of the problem, we can denote w.l.o.g
$$R_{opt}(r) = \begin{pmatrix} r & 0 \\ 0 & r \end{pmatrix}$$
for some $r>0$, and ask what value of $r$ would result in minimal square filtering errors of the KF.

\begin{lemma}[Optimization in the simplified lidar problem]
    \label{lemma:lidar_OKF}
    Assume that at a certain time-step in the simplified lidar problem, the state of the system is known to be normally-distributed with mean and covariance
    $$x_0=(x_1,x_2)^\top \qquad P_0=\begin{pmatrix} p & 0 \\ 0 & p \end{pmatrix}$$
    and that a new observation $z=(x_1+dx_1,x_2+dx_2)^\top$ is received.
    Denote by $x$ the point-estimate of the KF following a filtering step with respect to noise parameters of the form $R_{opt}(r)$.
    Then, $r=\frac{pr_0}{2p+r_0}$ minimizes the expected square error ($MSE$) of $x$.
    Furthermore, $(R_{opt})_{ii}=r<(R_{est})_{ii}$ (for $i\in \{1,2\}$), i.e., the optimal value of $r$ is smaller than the value obtained by noise estimation.
\end{lemma}
\begin{proof}
    As the true model of the observation noise is $R(\theta)$, by applying the KF update step (specified in Figure~\ref{fig:KF}) we obtain the new true mean of the system-state distribution:
    \begin{align*}
        x_{true} &= x_0 + P_0H^\top(HP_0H^\top+R)^{-1}(z-Hx_0) \\ &= x_0 + P_0(P_0+R)^{-1}(z-x_0) \\
        &= x_0 + \begin{pmatrix}
            \frac{r_0\sin^2\theta+p}{p+r_0} & -\frac{r_0\cos\theta\sin\theta}{p+r_0} \\
            -\frac{r_0\cos\theta\sin\theta}{p+r_0} & \frac{r_0\cos^2\theta+p}{p+r_0}
        \end{pmatrix} \begin{pmatrix}
            dx_1 \\ dx_2
        \end{pmatrix} \\ &= \begin{pmatrix}
            x_1 + \frac{(r_0\sin^2\theta+p)dx_1 - (r_0\cos\theta\sin\theta)dx_2}{p+r_0} \\
            x_2 + \frac{(r_0\cos^2\theta+p)dx_2 - (r_0\cos\theta\sin\theta)dx_1}{p+r_0}
        \end{pmatrix}
    \end{align*}
    where $\theta$ corresponds to the (unknown) true direction of the target.
    However, due to the restricted form of $R_{opt}(r)$ (which is independent of $\theta$), the actual estimate would be
    $$x=(x_1+\frac{p}{p+r}dx_1,x_2+\frac{p}{p+r}dx_2)^\top$$
    
    The expected error of the point-estimate $x$ can be decomposed into two terms: the bias of $x$, and the variance of the system-state distribution. More formally, we can write $MSE = MSE_{var}(P_{true}) + MSE_{bias}(x, x_{true})$, where we can only control the latter.
    Specifically, we denote $a(r)\coloneqq p/(p+r)$ and look for the value of $a$ that minimizes the $MSE$. We use the identities $\sin2\theta=2\cos\theta\sin\theta$ and (according to $F$ in Definition~\ref{def:simplified_lidar}) $E\left[dx_1^2\right]=E\left[dx_2^2\right]=q$.
    \begin{align*}
        MSE_{bias}&(x(a), x_{true}) = E||x-x_{true}||^2 \\
        =& E\left[
        \left((a-\frac{r_0\sin^2\theta+p}{p+r_0})dx_1 + \frac{r_0\sin(2\theta)/2}{p+r_0}dx_2\right)^2\right. \\
        &+ \left.\left((a-\frac{r_0\cos^2\theta+p}{p+r_0})dx_2 + \frac{r_0\sin(2\theta)/2}{p+r_0}dx_1\right)^2\right] \\
        =& E\left[
        dx_1^2\left( a^2 - 2a\frac{r_0\sin^2\theta+p}{p+r_0} + C_1 \right) \right. \\ &+ \frac{r_0^2\sin^2(2\theta)/4}{(p+r_0)^2}dx_2^2 + A_1dx_1dx_2 \\
        &+ dx_2^2\left( a^2 - 2a\frac{r_0\cos^2\theta+p}{p+r_0} + C_2 \right) \\ &+ \left. \frac{r_0^2\sin^2(2\theta)/4}{(p+r_0)^2}dx_1^2 + A_2dx_1dx_2
        \right] \\
        =& 2qa^2 - 2qa\frac{r_0+2p}{p+r_0} + q(C_1+C_2) + q\frac{r_0^2\sin^2(2\theta)/2}{(p+r_0)^2}
    \end{align*}
    where $C_{1,2}$ are independent of $a$, and $A_{1,2}$ are multiplied by $E[dx_1dx_2]=0$ and vanish. To minimize we calculate
    \begin{align*}
        0 &= \frac{\partial MSE_{bias}(x(a), x_{true})}{\partial a} = 4q\cdot a - 2q\frac{2p+r_0}{p+r_0}
    \end{align*}
    which gives us
    \begin{align*}
        a = \frac{p+r_0/2}{p+r_0}
    \end{align*}
    Note that the expression of $MSE_{bias}$ above clearly diverges as $|a|\rightarrow \infty$, hence the only critical point necessarily corresponds to a minimum of the $MSE$.
    Thus, the optimal $MSE$ is given when substituting the following $r$ in $R_{opt}$:
    \begin{align*}
        r &= p/a-p = \frac{p^2+pr_0 - (p^2+pr_0/2)}{p+r_0/2} = \frac{pr_0}{2p+r_0}
    \end{align*}
    
    For the last part of the lemma, we recall that $(R_{est})_{ii}=r_0/2$ and compare to $r$ directly:
    \begin{align*}
        (R_{est})_{ii} - (R_{opt})_{ii} &= r_0/2-r = \frac{(pr_0 + r_0^2/2) - pr_0}{2p+r_0} = \frac{r_0^2/2}{2p+r_0} > 0
    \end{align*}
\end{proof}

Lemma~\ref{lemma:lidar_OKF} shows that independently of the value of $x_0$ (i.e., independently of the current state of the system or the current direction of the target), the expected filtering error would benefit from reduction of the noise parameter to $(R_{opt})_{ii} = r < (R_{est})_{ii}$.
In particular, noise estimation is not an optimal strategy for parameters tuning in this problem.

\begin{figure}[!h]
\centering
\includegraphics[width=.4\linewidth]{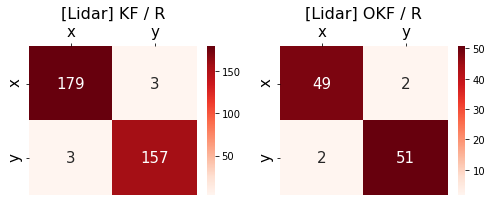}
\caption{\small The observation noise covariance $R$ obtained by noise estimation (KF) and by optimization (OKF), for the problem of state-estimation from lidar measurements (Section~\ref{sec:lidar}).}
\label{fig:lidar_noise}
\end{figure}

\textbf{Empirical results:}
The simplified lidar problem can be seen as a simplification of the problem experimented in Section~\ref{sec:lidar}, with a degenerated motion model.
In the experiments, OKF reduced the $MSE$ by 14.9\%, and the optimized values of $R$ were indeed smaller than the estimated observation noise, as shown in Figure~\ref{fig:lidar_noise}.

In summary, in this section we introduced a simplified lidar problem. While at first glance the problem appears to follow the KF assumptions, the i.i.d property of the noise is in fact violated by the spherical nature of the sensor.
Consequentially, the optimal values of $R$ do not correspond anymore to the noise covariance, but rather should be reduced. In other words, {\bf the non-i.i.d nature of the noise reduces the \textit{effective} noise}.
This theoretic result is consistent with the experiments presented in Section~\ref{sec:lidar}.



\section{Preliminaries: Recurrent Neural Networks}
\label{sec:preliminaries_extended}


\textit{Neural networks} (NN) are parametric functions, usually constructed as a sequence of matrix-multiplications with some non-linear differentiable transformation between them. NNs are known to be able to approximate complicated functions, given that the right parameters are chosen. Optimization of the parameters of NNs is a field of vast research for decades, and usually relies on gradient-based methods, that is, calculating the errors of the NN with relation to some training-data of inputs and their desired outputs, deriving the errors with respect to the network's parameters, and moving the parameters against the direction of the gradient.

\textit{Recurrent neural networks} (RNN)~\citep{RNN} are NN that are intended to be iteratively fed with sequential data samples, and that pass information (the \textit{hidden state}) over iterations. Every iteration, the hidden state is fed to the next copy of the network as part of its input, along with the new data sample.
\textit{Long Short Term Memory} (LSTM)~\citep{LSTM} is an architecture of RNN that is particularly popular due to the linear flow of the hidden state over iterations, which allows to capture memory for relatively long term.
The parameters of a RNN are usually optimized in a supervised manner with respect to a training dataset of input-output pairs.

\section{Generalization Tests: Training and Testing on Different Benchmarks}
\label{sec:generalization}

In the case-study of Section~\ref{sec:OKF}, we demonstrate the robustness of an Optimized KF (OKF) over a variety of benchmarks representing different tracking scenarios -- where in every benchmark, OKF (Method~\ref{method:OKF}) obtained lower estimation errors than a standard KF (Method~\ref{method:KF}), {\bf over an out-of-sample test data}.
This demonstrates that OKF did not overfit the training data, but still relies on representativeness of the training data, i.e., on the assumption that the training data and the test data are taken from the same distribution.

A stronger generalization ability is demonstrated in Section~\ref{sec:NKF}, where OKF beats both KF and NKF not only on out-of-sample test data with the same settings, but also {\bf on a test data with different settings}. Specifically, we consider changes of factors 0.5 and 2 in the scale of the targets acceleration. This essentially changes the motion of the targets, including turns sharpness, speed changes and typical scale of speed. In terms of the linear motion model of KF, the acceleration corresponds to the motion noise $Q$, hence we essentially changed the noise after it had been learned, which poses a significant generalization challenge.
Thus, the results of Section~\ref{sec:NKF} provide a significant evidence for the robustness of OKF to certain distributional shifts.

\begin{figure}
\centering
\begin{subfigure}{\textwidth}
    \includegraphics[width=\linewidth]{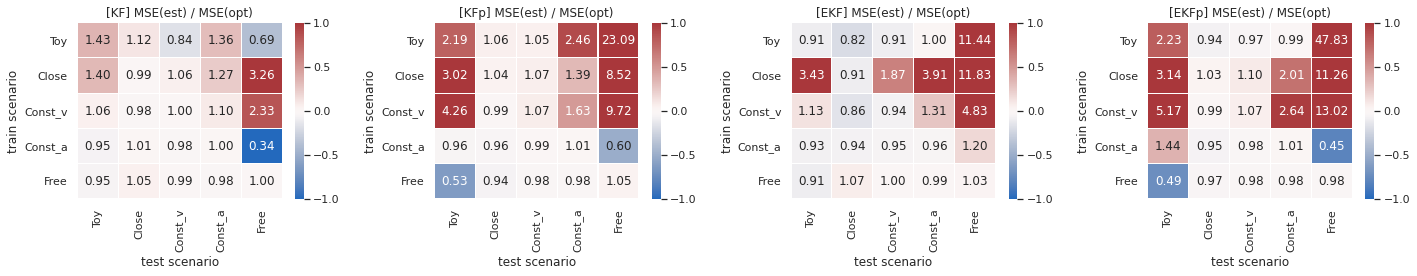}
    \caption{\small $MSE\_ratio = MSE(KF) / MSE(OKF)$ for every KF-baseline (KF,KFp,EKF,EKFp defiend in Section~\ref{sec:OKF}), and for every pair of train-scenario and test-scenario. The colormap is logarithmic ($\sim log(MSE\_ratio)$), where red values represent advantage to OKF ($MSE\_ratio>1$).}
    \label{fig:cross_scenarios_detailed}
\end{subfigure}
\begin{subfigure}{.5\textwidth}
    \includegraphics[width=\linewidth]{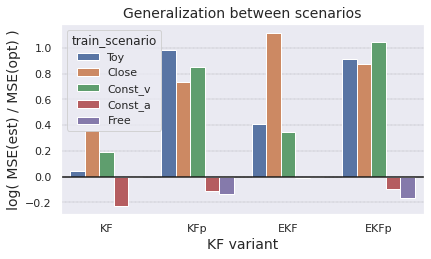}
    \caption{\small For every train-scenario, $MSE\_ratio$ is averaged over all the test-scenarios and is shown in a logarithmic scale. Positive values indicate advantage to OKF.}
    \label{fig:cross_scenarios_summary}
\end{subfigure}
\caption{\small Generalization tests: optimization vs. noise estimation under distributional shifts.}
\label{fig:cross_scenarios}
\end{figure}

In this section we present a yet stronger evidence for the robustness of OKF -- not over a parametric distributional shift, but {\bf over entirely different benchmarks}. Specifically, we consider the 5 benchmarks (or scenarios) of Section~\ref{sec:OKF}. For every pair (train-scenario, test-scenario), we train both KF and OKF on data of the train-scenario, then test them on data of the test-scenario.
For every such pair of scenarios, we measure the generalization advantage of OKF over KF through $MSE\_ratio = MSE(KF) / MSE(OKF)$ (where $MSE\_ratio>1$ indicates positive advantage). To measure the total generalization advantage of a model trained on a certain scenario, we calculate the geometric mean of $MSE\_ratio$ over all the test-scenarios (or equivalently, the standard mean over the logs of the ratios). Note that the logarithmic scale is indeed more natural for a symmetric view of the ratio of the two MSE scores.

This test is quite noisy, since a model optimized for a certain scenario may legitimately be inferior in other scenarios. Yet, considering all the results together in Figure~\ref{fig:cross_scenarios}, it is evident that optimization of KF yields more robust models in most cases, and even when it \textit{is} inferior to noise-estimation, it does not lose by a large margin.


\section{Sensitivity of KF Optimization to Train Dataset Size}
\label{sec:train_size}

Each benchmark in the case-study of Section~\ref{sec:OKF} has 1500 targets in its train data.
Training from these datasets demonstrated significant advantage to optimization of KF over noise-estimation.
However, one may argue that on smaller datasets, optimization may struggle harder than noise-estimation, e.g., due to numeric instability with little data. Furthermore, the standard optimization procedure "wastes" a portion of the train data as a validation set, which does not contribute directly to the training, and thus may increase the sensitivity to the amount of data.

\begin{figure}
\centering
    \includegraphics[width=0.85\linewidth]{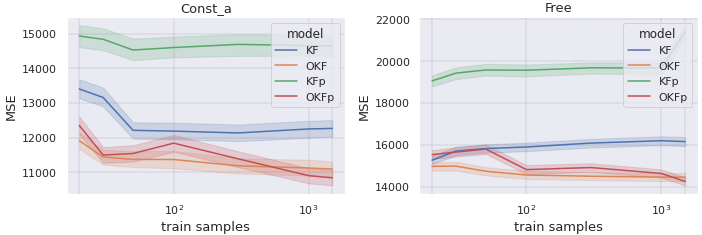}
\caption{\small The advantage of OKF over KF remains positive -- although sometimes smaller -- even when the train data reduces to as few as 20 targets. The shadowed areas correspond to 95\% confidence-intervals.}
\label{fig:train_size}
\end{figure}

To test the sensitivity of KF optimization to the amount of data, we repeated some of the tests of Section~\ref{sec:OKF} using smaller subsets of the train datasets -- beginning from as few as 20 targets per dataset.
Figure~\ref{fig:train_size} demonstrates that the advantage of OKF over KF holds consistently over the various sizes of the train datasets, although it indeed increases with the amount of data.
In fact, in the Free Motion benchmark, the state-estimation errors of the standard KF monotonously \textit{increase} with the amount of data.

\section{Neural Kalman Filter: Extended Experiments}
\label{sec:NKF_extended}

Section~\ref{sec:NKF} demonstrates that even if a neural-network (NN) based model provides more accurate predictions than a standard KF -- its advantage may be entirely vanished when compared to an optimized KF instead.
It is shown for the specific NKF model in a free-motion benchmark in the Doppler radar problem.
This is sufficient to demonstrate the error in the common methodology that compares optimized NN-based models to non-optimized variants of KF.

To show that these results are not exclusive to the experiments of Section~\ref{sec:NKF}, we present extended experiments in this section.
Specifically, we consider 2 different benchmarks -- the free-motion benchmark of Section~\ref{sec:NKF} and Const\_a benchmark of Section~\ref{sec:OKF} (in which the targets may have acceleration but not turns). We also consider 3 different neural models:
\begin{itemize}
    \item Predicted-acceleration KF (aKF): a variant of NKF without dynamic prediction of the covariance matrices $Q$ and $R$ in every step.
    \item Neural KF (NKF): the model used in Section~\ref{sec:NKF} and shown in Figure~\ref{fig:NKF_diagram}.
    \item Neural KF with H-prediction (NKFH): a variant of NKF that also predicts the observation model $H$ dynamically in every step.
\end{itemize}

For each benchmark and each model, we trained the model on train data with a certain range of targets acceleration (note that acceleration affects both speed changes and turns sharpness), and tested it on targets with three different acceleration ranges -- the original one, a smaller one and a larger one. That is, some of the test datasets correspond to distributional shifts.
For each model we train two variants -- one with Cartesian representation of the observation noise $R$, and one with spherical representation (as in the baselines of Section~\ref{sec:OKF}) -- and we select the best among the two variants using the MSE scores on the validation data (which is a portion of the data assigned for training).

We measure the results using (1) the Mean Square Error (MSE) of the estimated location after each update step; and (2) the Negative-Log-Likelihood (NLL) of the true system-state with relation to the estimated state-distribution, after the prediction step. Note that NLL is important for matching observations to targets in the multi-target tracking problem (which is out of the scope of this work).

\begin{figure}[!h]
\centering
\begin{subfigure}{\textwidth}
    \includegraphics[width=\linewidth]{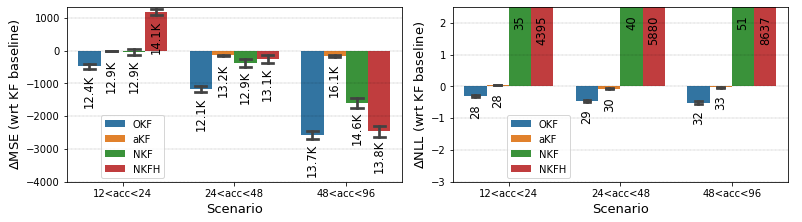}
    \caption{\small Free-motion benchmark}
    \label{fig:NKF_extended_free}
\end{subfigure}
\begin{subfigure}{\textwidth}
    \includegraphics[width=\linewidth]{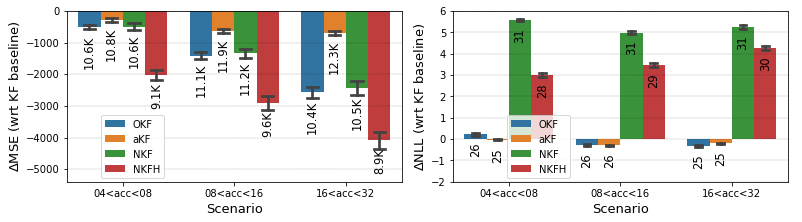}
    \caption{\small Const\_a benchmark (no turns)}
    \label{fig:NKF_extended_consta}
\end{subfigure}
\caption{\small The \textit{relative} MSE and NLL results of various models in comparison to the standard KF model. The textual labels specify the \textit{absolute} MSE and NLL. Note that certain bars of NLL are of entirely different scale and thus are cropped in the figure (their values can be seen in the labels).
In each benchmark, the models were trained with relation to MSE loss, on train data of the middle acceleration-range: the two other acceleration ranges in each benchmark correspond to generalization over distributional shifts.}
\label{fig:NKF_extended}
\end{figure}

Figure~\ref{fig:NKF_extended_free} shows that in the free-motion benchmark, all the 3 neural models improve the MSE in comparison to a standard KF, but lose to an optimized one (OKF).
Furthermore, while OKF has the best NLL scores, the more complicated models NKF and NKFH increase the NLL in orders of magnitude. This issue may be handled by explicitly optimize the NLL in addition to the MSE, but such multi-loss optimization is out of the scope of this work.
Note that the instability of NKFH is expressed in poor generalization to lower accelerations in addition to the extremely high NLL score.

Figure~\ref{fig:NKF_extended_free} shows that in Const\_a benchmark, all the 3 neural models improve the MSE in comparison to a standard KF, but only NKFH improves in comparison to OKF as well.
Note that while NKFH does better in this case than in the free-motion benchmark, it still suffers from very high NLL.

In summary, by comparing the optimized neural models to the standard KF model, all the 3 models would be found superior to the KF in both benchmarks in terms of MSE. However, when shifting the baseline to an \textit{optimized} KF, we find that the neural aKF and NKF are in fact inferior to OKF, and that the comparison between NKFH and OKF depends on the selected benchmark and metric.

Of course, our results do not imply that neural-networks in general cannot be superior to a KF: only that when comparing the two, one must optimize the KF similarly to the neural model to avoid misleading experimental results. As discussed in Section~\ref{sec:related_work}, the wrong methodology of using a non-optimized KF baseline is very common in the literature.


\section{On the Guarantees of Gradient-based Optimization}
\label{sec:GD_limitations}

Method~\ref{method:OKF} can rely on any gradient-based optimization algorithm, such as SGD or Adam~\citep{adam} (the latter is used in this work).
Such optimization algorithms have achieved remarkable results in a variety of optimization problems in the recent years. This includes impressive results in non-convex problems where local-minima exist~\citep{adam_revisited}, as well as generalization to unseen data~\citep{GD_generalization}.

Compared to the problems where algorithms like Adam are often applied (e.g., language models with more than 100M parameters~\citep{bert}), typical filtering problems such as the ones experimented in this work are arguably small and simple to optimize.
Thus, the robust results of Method~\ref{method:OKF} -- namely, the consistent advantage over Method~\ref{method:KF} in \textit{all} the experiments in this work -- should not come as a surprise.

Having said that, the \textit{theoretical} guarantees of the gradient-based algorithms are limited for non-convex problems.
Note that the dependence of our loss function (filtering errors) on the model parameters (after the Cholesky parameterization) is indeed not convex.
In such settings, convergence to a local optimum of the loss \textit{is} in fact guaranteed: Adam, for example, is guaranteed to converge given a twice-differentiable loss function~\citep{adam_convergence}.
Stronger convergence guarantees exist for the SGD algorithm under certain conditions~\citep{SGD_convergence}, yet in general, gradient-based optimization algorithms do not necessarily converge to the global optimum of the loss.
Furthermore, the global optimum is not necessarily unique.

Note that in all the problems experimented in this work, the standard Method~\ref{method:KF} (noise-estimation) has no optimality guarantee either, as the KF assumptions do not hold.
Indeed, the experiments demonstrate its sub-optimality, and Appendices~\ref{sec:toy_analysis},\ref{sec:theory_noniid} prove it theoretically for a sample of problems.


\section{Detailed Results}
\label{sec:detailed_results}

\begin{figure}[h]
\centering
\begin{subfigure}{.8\textwidth}
  \centering
  \includegraphics[width=1.\linewidth]{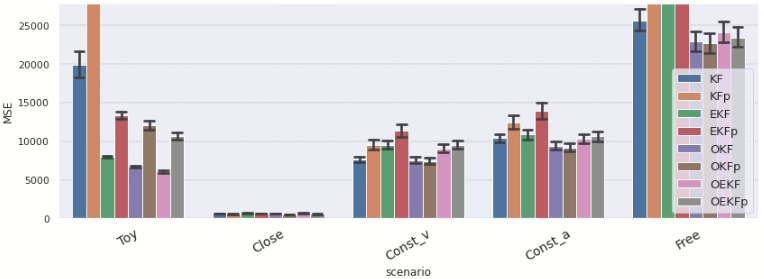}
  \caption{}
  \label{fig:res_all_KF}
\end{subfigure} \\
\begin{subfigure}{.45\textwidth}
  \centering
  \includegraphics[width=1.\linewidth]{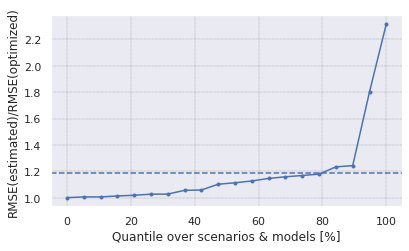}
  \caption{}
  \label{fig:res_relative}
\end{subfigure}
\caption{\footnotesize (a) Errors of different tracking models over different benchmarks. This is a different visualization of the results of Table~\ref{tab:res_OKF}. The error-bars correspond to confidence intervals of 95\%. Note that each of the optimized models (the last four) consistently yields lower errors than the corresponding estimated model (among the first four) -- in most cases with high statistical significance. (b) The RMSE ratio between the estimated models and the optimized ones, over all the benchmarks and designs discussed in Section~\ref{sec:OKF}. Note that all the ratios are above 1, i.e., all the models in all the benchmarks had smaller errors when tuned by optimization. The horizontal line represents the average ratio.}
\label{fig:res_OKF_detailed}
\end{figure}

\begin{figure}[!h]
\centering
\begin{subfigure}{.24\textwidth}
  \centering
  \includegraphics[width=1.\linewidth]{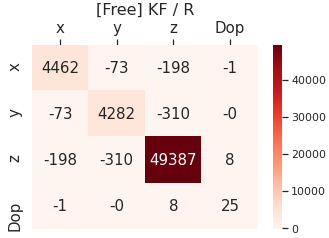}
  \caption{\footnotesize $R$ (estimated)}
  \label{fig:noise_R_KF}
\end{subfigure}
\begin{subfigure}{.24\textwidth}
  \centering
  \includegraphics[width=1.\linewidth]{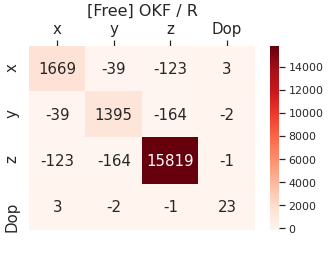}
  \caption{\footnotesize $R$ (optimized)}
  \label{fig:noise_R_OKF}
\end{subfigure}
\begin{subfigure}{.24\textwidth}
  \centering
  \includegraphics[width=1.\linewidth]{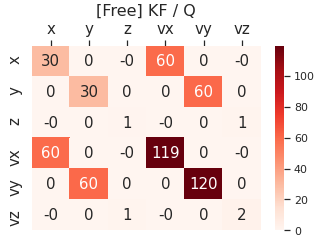}
  \caption{\footnotesize $Q$ (estimated)}
  \label{fig:noise_Q_KF}
\end{subfigure}
\begin{subfigure}{.24\textwidth}
  \centering
  \includegraphics[width=1.\linewidth]{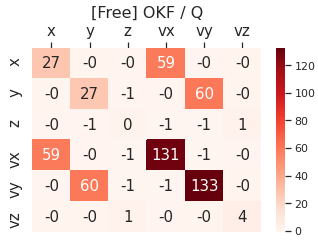}
  \caption{\footnotesize $Q$ (optimized)}
  \label{fig:noise_Q_OKF}
\end{subfigure}
\caption{\footnotesize The observation noise ($R$) and prediction noise ($Q$) matrices obtained in a (Cartesian) KF by noise estimation and by optimization with relation to $MSE$, based on the dataset of the Free-motion benchmark.}
\label{fig:noise_params}
\end{figure}

\begin{figure}[!h]
\centering
\begin{subfigure}{.32\textwidth}
  \centering
  \includegraphics[width=1.\linewidth]{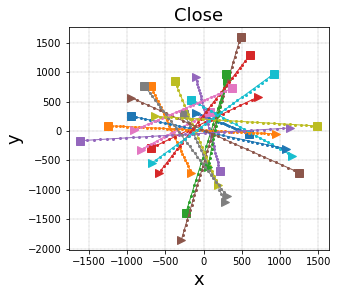}
\end{subfigure}
\begin{subfigure}{.32\textwidth}
  \centering
  \includegraphics[width=1.\linewidth]{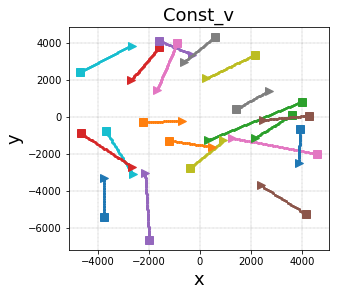}
\end{subfigure} 
\begin{subfigure}{.32\textwidth}
  \centering
  \includegraphics[width=1.\linewidth]{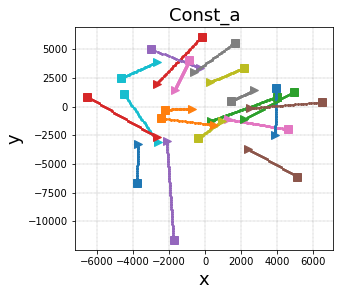}
\end{subfigure}
\caption{\footnotesize Samples of targets trajectories in the various benchmarks described in Section~\ref{sec:OKF}, projected onto the horizontal plane. This is an extension of Figures~\ref{fig:trajectories_toy},\ref{fig:trajectories_free} for the rest of the benchmarks.}
\label{fig:all_trajectories}
\end{figure}

\begin{figure}[!h]
  \centering
  \includegraphics[width=1.\linewidth]{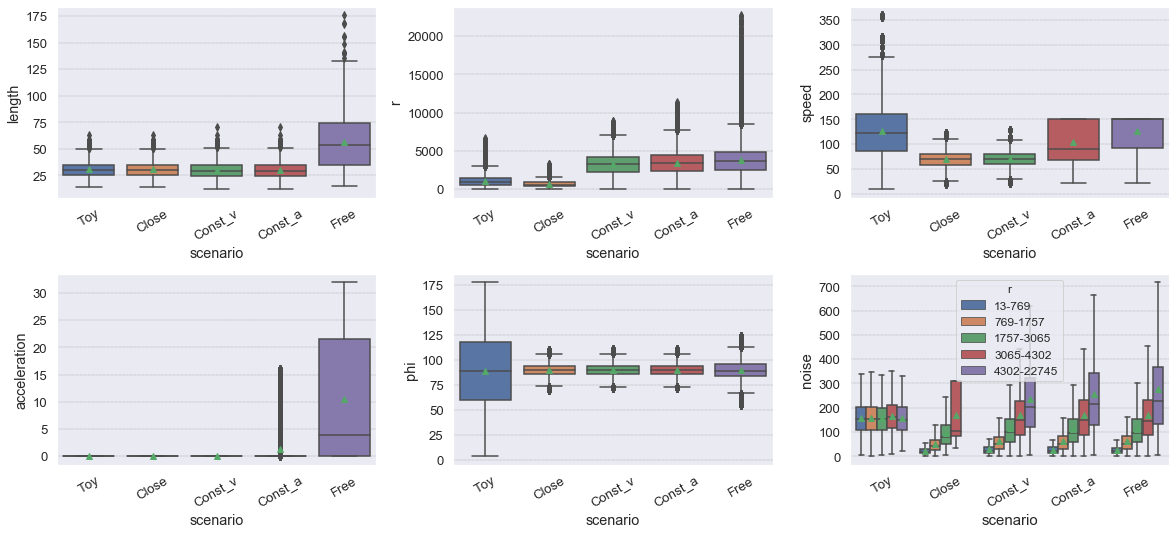}
    \caption{\footnotesize Descriptive statistics of the various benchmarks of Section~\ref{sec:OKF}: duration of targets trajectories; distance from the radar; targets speed; targets acceleration; motion direction (90 degrees correspond to horizontal motion); and observation errors vs. distance from radar.}
    \label{fig:benchmarks_EDA}
\end{figure}

Figures~\ref{fig:res_OKF_detailed}-\ref{fig:benchmarks_EDA} below provide additional details about the experiments discussed in Sections~\ref{sec:OKF},\ref{sec:NKF}.


\end{document}